\documentclass[twocolumn]{gewu}

\usepackage{wasysym}

\usepackage[toc,page,header]{appendix}

\usepackage{minitoc}
\usepackage{microtype}      % microtypography
\usepackage[dvipsnames,table]{xcolor}         % colors

\usepackage{hyperref}
\usepackage{url}
\usepackage{booktabs}
\usepackage{multirow}
\usepackage{subcaption}
\usepackage{caption}
\usepackage{amssymb}
\usepackage{graphicx}
\usepackage{amsmath}
\usepackage{wrapfig}
\usepackage{enumitem}
\usepackage{arydshln} 
\usepackage{gensymb}
\usepackage{makecell}
\usepackage{pifont}

\usepackage{algorithm}
\usepackage{ragged2e}
\usepackage{algpseudocode}
\usepackage{amsmath}
\usepackage{amssymb}
\usepackage{dsfont}
\definecolor{ModelGreen}{RGB}{213,232,212}

\title{APPO: Attention-guided Perception Policy Optimization \\ for Video Reasoning}

\author[1,2,*]{Henghui Du}
\author[2,\textnormal{\Letter}]{Chang Zhou}
\author[2,\textnormal{\Letter}]{Xi Chen}
\author[1,\textnormal{\Letter}]{Di Hu}

\affiliation[1]{Gaoling School of Artificial Intelligence, Renmin University of China, Beijing}
\affiliation[2]{AI Technology Center, Online Video Business Unit, Tencent PCG}

\contribution[*]{Work done at Tencent PCG}
\contribution[\textnormal{\Letter}]{ Co-Corresponding author}

\abstract{
    Complex video reasoning, actually, relies excessively on fine-grained perception rather than on expert (\emph{e.g.}, Ph.D, Science)-level reasoning.
Through extensive empirical observation, we have recognized the critical impact of perception.
In particular, when perception ability is almost fixed, enhancing reasoning from Qwen3-8B to OpenAI-o3 yields only $0.7\%$ performance improvement. Conversely, even minimal change in perception model scale (from 7B to 32B) boosts performance by $1.4\%$, indicating enhancing perception, rather than reasoning, is more critical to improve performance.
Therefore, exploring how to enhance perception ability through reasoning without the need for expensive fine-grained annotation information is worthwhile.
To achieve this goal, we specially propose \textbf{APPO}, the \textbf{A}ttention-guided \textbf{P}erception \textbf{P}olicy \textbf{O}ptimization algorithm that leverages token-level dense rewards to improve model's fine-grained perception.
The core idea behind APPO is to optimize those tokens from different responses that primarily focus on the same crucial video frame (called intra-group perception tokens).
Experimental results on diverse video benchmarks and models with different scales ($3$/$7$B) demonstrate APPO consistently outperforms GRPO and DAPO ($0.5\%\sim4\%$). 
We hope our work provides a promising approach to effectively enhance model's perception abilities through reasoning in a low-cost manner, serving diverse scenarios and demands.
}

\MyEmail{\textsuperscript{1}\{cserdu, dihu\}@ruc.edu.cn, \textsuperscript{2}\{cserdu, chanzhou, jasonxchen\}@tencent.com}
\checkdata[Project Page]{\url{https://gewu-lab.github.io/APPO}}
\checkdata[Code]{\url{https://github.com/GeWu-Lab/APPO}}
\date{\today}

\begin{document}
% \maketitle

%不需要目录就注释掉 注意目录不要和第一页放在一块 要有\newpage
%\newpage
%\tableofcontents
%\newpage

\twocolumn[{
\renewcommand\twocolumn[1][]{#1}
\maketitle
\vspace{-2.25em}
\centering
\captionsetup{type=figure}\includegraphics[width=\linewidth]{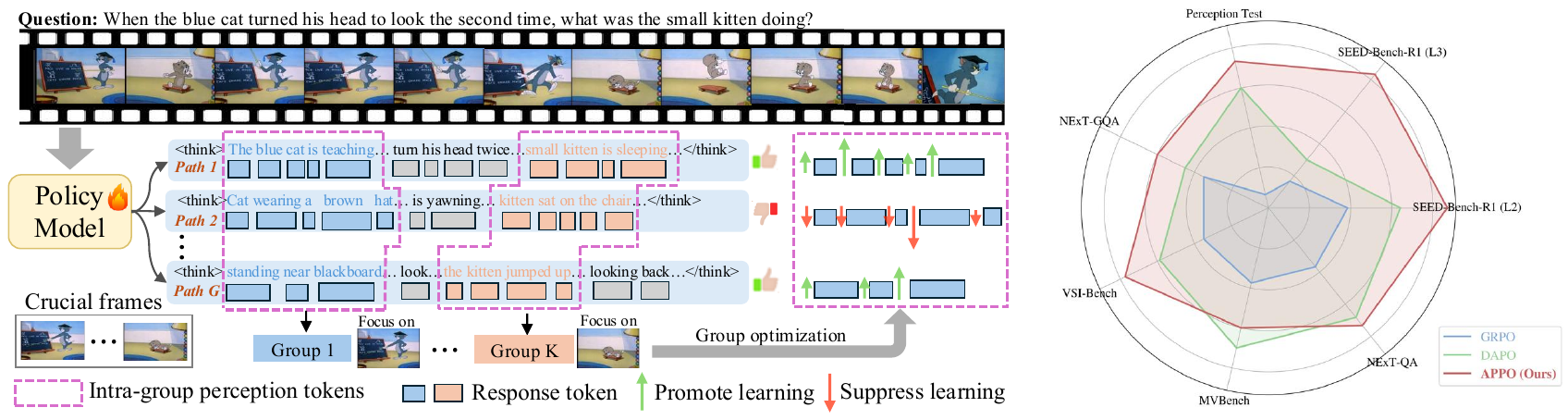}
\vspace{-2.25em}
\captionof{figure}{
We present \textbf{APPO}, the \textbf{A}ttention-guided \textbf{P}erception \textbf{P}olicy \textbf{O}ptimization algorithm that enhances model's fine-grained perception ability through reasoning.
The core idea behind APPO is to optimize those tokens from different responses that primarily focus on the same crucial video frames (called \emph{intra-group perception tokens}), resulting in fine-grained token level reward signals. 
\textbf{Left:} The illustration of APPO algorithm. The intra-group perception tokens are defined as those tokens from different responses that primarily focus on the same crucial video frame. The perception tokens within each group are optimized with different learning intensities.
\textbf{Right:} Experimental results on multiple video benchmarks demonstrate APPO achieves overall performance improvement compared with GRPO and DAPO.
}
\label{fig:teaser}
\vspace{1em}
}]

\section{Introduction}
\label{sec:intro}
Reinforcement Learning with Verifiable Rewards (RLVR) has proven to be a highly effective strategy for endowing Large Language Models (LLMs) with powerful reasoning abilities to solve complex tasks~\cite{jaech2024openai, guo2025deepseek}. DeepSeek-R1~\cite{guo2025deepseek} employs verifiable signals, such as structured thinking formats and final answer accuracy to optimize models through Group Relative Policy Optimization (GRPO) algorithm. Subsequent works, such as DAPO~\cite{yu2025dapo} and GSPO~\cite{zheng2025group} \emph{etc.}, have consistently achieved superior performance improvements compared to Supervised Fine-Tuning (SFT).
Motivated by this success, a growing body of works~\cite{feng2025video, park2025deepvideo, grpogrpo, dang2025reinforcing} have explored applying RLVR to video scenarios, in hopes of improving the video reasoning abilities of MLLMs. By specially designing sparse outcome reward signals for different types of tasks, such as IoU of bounding box and timestamps, \emph{etc.}, they have yielded remarkable success, particularly in terms of generalization ability.

% by specially designing reward signals for different types of tasks to optimize model, such as IoU of bounding box and timestamps, \emph{etc.}.

% In this situation, are optimization algorithms specifically tailored for text tasks still applicable in video scenarios?

\begin{figure*}[!t]
    \centering
    \begin{subfigure}[b]{0.32\textwidth}
        \centering
        \includegraphics[width=\textwidth]{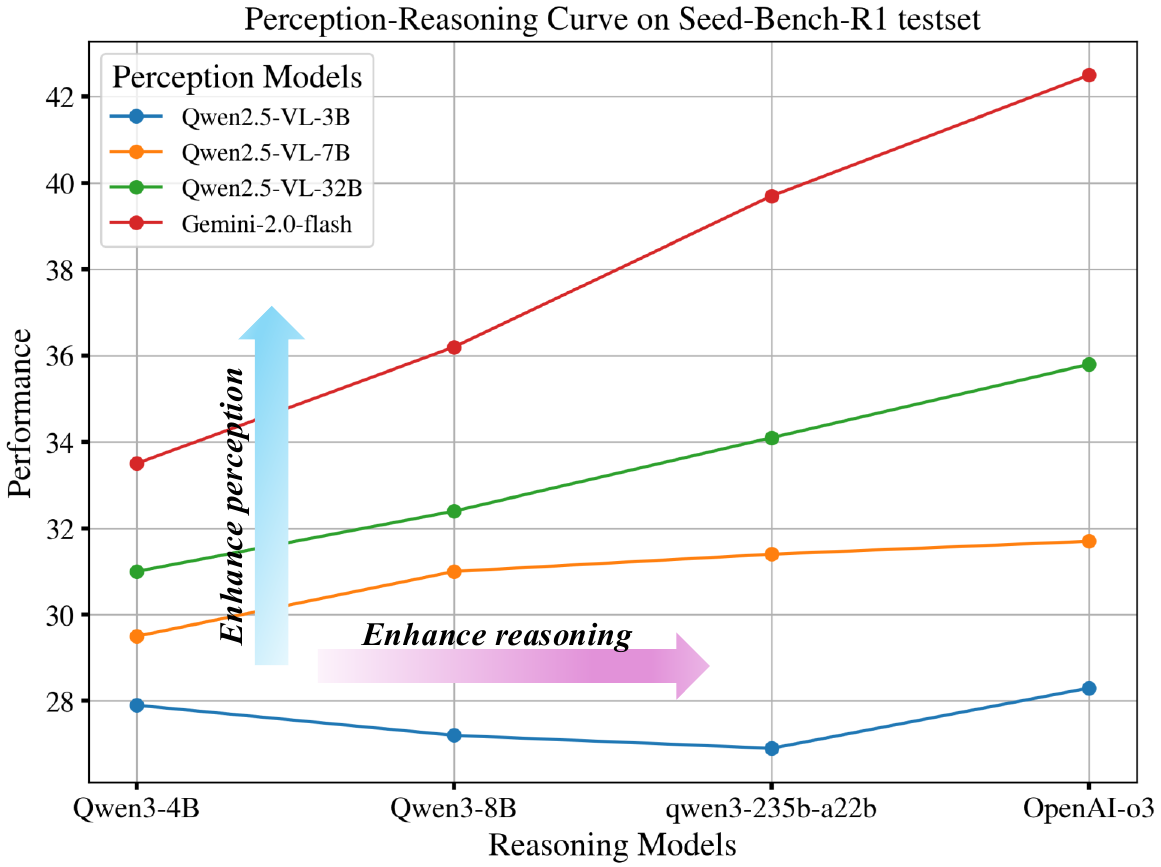}
        \caption{SEED-Bench-R1 benchmark.}
        \label{fig:toy-exp-seed-bench-r1}
    \end{subfigure}
    \hfill
    \begin{subfigure}[b]{0.32\textwidth}
        \centering
        \includegraphics[width=\textwidth]{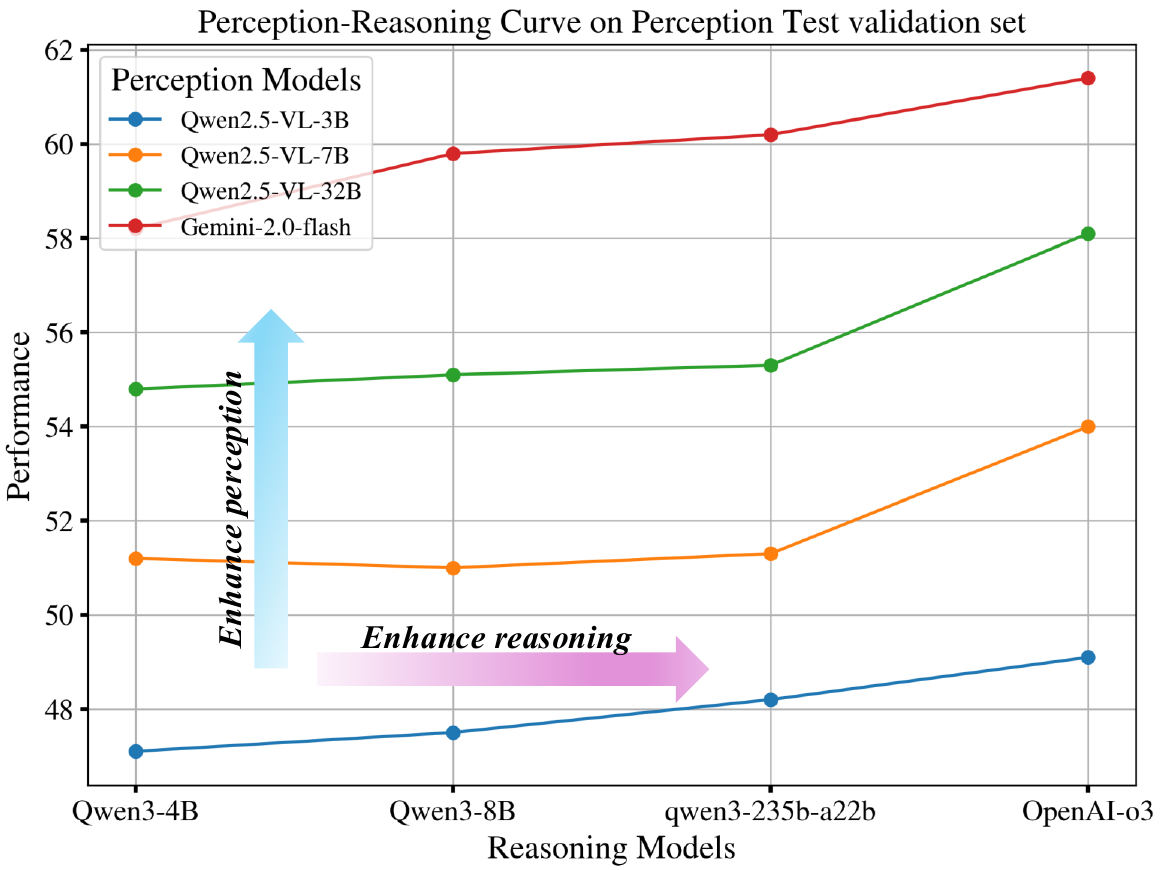}
        \caption{Perception Test benchmark.}
       \label{fig:toy-exp-percepton-test}
    \end{subfigure}
    \hfill
    \begin{subfigure}[b]{0.32\textwidth}
        \centering
        \includegraphics[width=\textwidth]{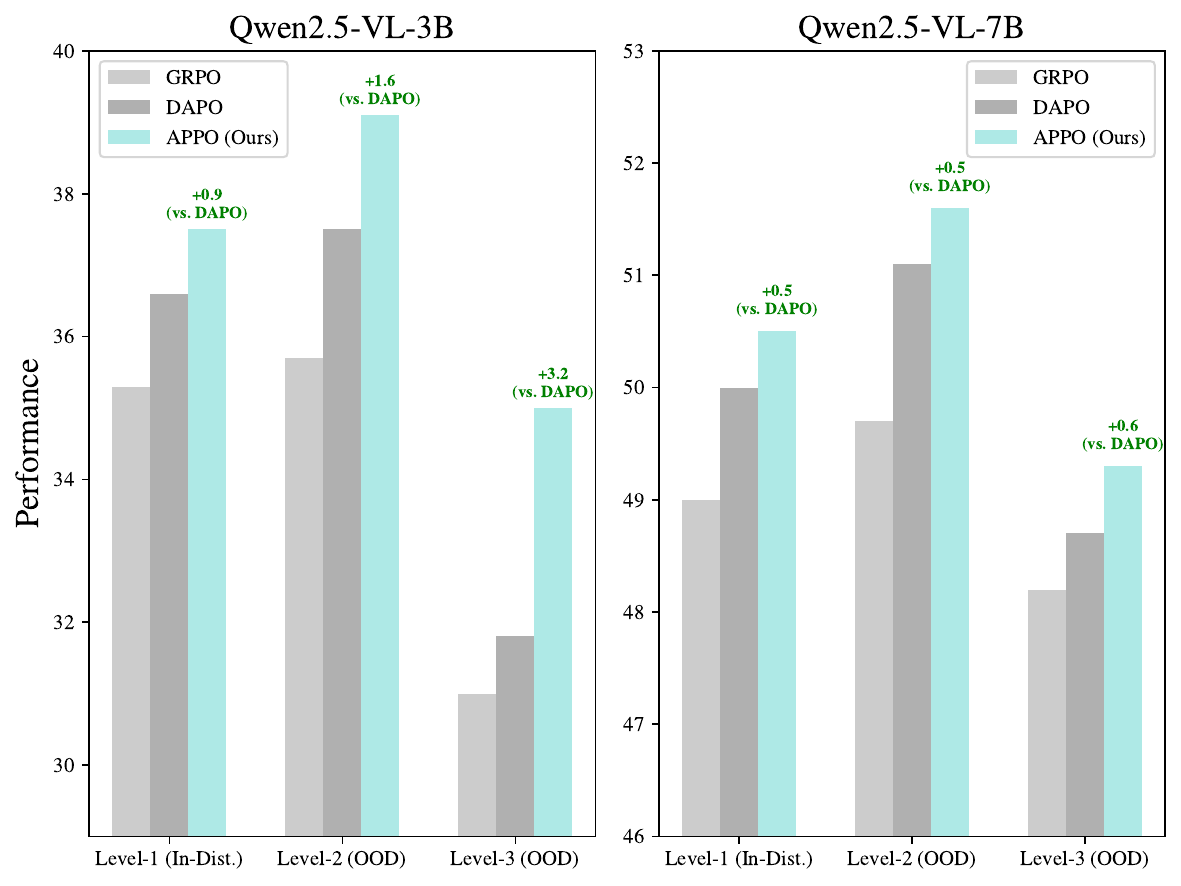}
        \caption{Performance on SEED-Bench-R1.}
        \label{fig:performance_cmp}
    \end{subfigure}
    \vspace{-0.8em}
    \caption{The \textbf{Perception-Reasoning curves} on SEED-Bench-R1~\cite{chen2025exploring} and Perception-Test~\cite{patraucean2023perception} benchmarks, quantifying the impact of perception \textit{vs.} reasoning ability on overall performance.
    % which are specially constructed to evaluate model's perception and reasoning skills. 
    % that quantify the impact of perception vs. reasoning on overall model performance.
    Each point in the curve represents the performance achieved by combining specific perception and reasoning ability.
    In particular, we first prompted four perception models with progressively enhanced abilities (including Qwen2.5-VL-3/7/32B~\cite{bai2025qwen2} and Gemini-2.0-flash~\cite{comanici2025gemini}) to describe video content in detail. Subsequently, the other four reasoning models with varying capabilities (including Qwen3-4/8B, Qwen3-235-A22B-thinking~\cite{yang2025qwen3}, and OpenAI-o3~\cite{jaech2024openai}) were used to think and answer questions based on the descriptions provided by each perception model, respectively, yielding $4 \times 4$ cross-combination results.
    % The final performance is evaluated on SEED-Bench-R1~\cite{chen2025exploring} and Perception-Test~\cite{patraucean2023perception}, which are specially constructed to evaluate model's perception and reasoning skills. 
    (a) For SEED-Bench-R1 benchmark, we evaluate on $2K$ Level-1 samples. (b) For Perception Test benchmark, we randomly select $1K$ samples from different videos for evaluation. (c) The performance comparison of GRPO, DAPO and our APPO on SEED-Bench-R1 benchmark across different scales models, demonstrating the significant improvements brought by enhanced perception.
 }
    \label{fig:toy-exp}
    \vspace{-1.5em}
\end{figure*}

% The specific prompt templates for perception and reasoning models are provided in the \emph{supplementary materials}.

Despite significant progress in most prior efforts, they have primarily focused on improving data quality~\cite{li2025truth, liang2025learning} or rewards design~\cite{wang2025time, ouyang2025spacer}.
It should be noted that different from text-based tasks, complex video reasoning involves \emph{fine-grained perception} and multi-step reasoning, both of which are essential and extremely important.
% It is worth noting that different from text-based tasks, which primarily focus on reasoning, complex video reasoning tasks additionally require accurate perception of video content.
% Intuitively, if the model perceives incorrect information or misses important details, it will struggle to reason out the correct outcomes.
Intuitively, perception is the foundation and premise for reasoning.
For example, as shown in Fig.~\ref{fig:teaser}, if the model fails to notice the critical behavior of the small kitten sleeping, or confuses the sequence of blue cat turning its head twice, it would be very difficult to reason out the correct answer. 
While it might be solved based on contextual information, such as the small kitten yawning and stretching, this requires stronger reasoning and self-correction capabilities, which increases the burden on the model.
Therefore, this situation presents us two fundamental research question: 
% \textbf{(1)} \emph{Does the optimization algorithm tailored for purely textual domains still be applicable in complex video scenarios?} \textbf{(2)} \emph{If not, how can we design a new optimization algorithm \textbf{tailored for video reasoning tasks} to bridge this gap?}
\textbf{(1)} \emph{For video reasoning, what is the key to improve model's performance, enhancing perception or reasoning?} \textbf{(2)} \emph{If perception, how to design a perception optimization strategy to effectively enhance fine-grained perception without relying on expensive fine-grained annotation or additional reward models?}
% bridge this gap by designing a new optimization algorim \textbf{tailored for video reasoning tasks}?}

% Intuitively, if the model perceives incorrect information or misses important details, it will struggle to reason out the correct outcomes. 
% To investigate the first question, we conducted a comprehensive experiment on two video benchmarks (SEED-Bench-R1~\cite{chen2025exploring} and Perception Test~\cite{patraucean2023perception}) that evaluate MLLMs's perception and reasoning abilities.

% To investigate the first question, we conducted a comprehensive experiment to explore the impact of perception and reasoning capabilities on model's performance.
% In particular, we first prompted four perception models with progressively enhanced perception abilities (including Qwen2.5-VL-3/7/32B~\cite{bai2025qwen2} and Gemini-2.0-flash~\cite{comanici2025gemini}) to describe video content in detail. Subsequently, the other four reasoning models with varying reasoning capabilities (including Qwen3-4B, Qwen3-8B, Qwen3-235-A22B-thinking~\cite{yang2025qwen3}, and OpenAI-o3~\cite{jaech2024openai}) were used to think and answer questions based on the descriptions provided by each perception model, respectively.
% We evaluated the final performance on two benchmarks: SEED-Bench-R1~\cite{chen2025exploring} and Perception-Test~\cite{patraucean2023perception}, which are constructed to evaluate model's perception and reasoning skills.
% As a result, the \textbf{Perception-Reasoning curves} on two benchmarks are shown in Fig.~\ref{fig:toy-exp}.

To investigate the first question, we employ an innovative divide-and-conquer strategy to quantify the impact of perception \emph{vs.} reasoning on overall performance.
% to decouple multimodal reasoning into perception and reasoning stages, quantifying the impact of perception \emph{vs.} reasoning on overall performance.
% Specifically, we modularize and decouple the capabilities of "perception" and "reasoning," using a cross-combination (4x4 full factorial/orthogonal design) to form a grid-based evaluation, and plot the perception-reasoning performance curve.
Specifically, we modularize the capabilities of perception and reasoning, leveraging four perception models and four reasoning models with progressively enhanced abilities.
By controlling variables, they are cross-combined to form a grid-based evaluation.
As a result, the \textbf{Perception-Reasoning curves} on two benchmarks are shown in Fig.~\ref{fig:toy-exp}.
It can be easily observed that: (1) when one capability (either perception or reasoning) is fixed, enhancing the other leads to an overall trend of improved performance; (2) enhancing perception ability results in a more remarkable improvement than enhancing reasoning. For example, on SEED-Bench-R1 benchmark, when the perception model is fixed as Qwen2.5-VL-7B, changing the reasoning model from Qwen3-8B to OpenAI-o3 only results in $0.7\%$ improvement; conversely, changing the perception model from Qwen2.5-VL-7B to Qwen-2.5-VL-32B, even this minimal change in model scale can enhance the performance of the Qwen3-8B model by $1.4\%$.
This suggests that in complex video scenarios, enhancing perception ability, rather than reasoning, is more important to improve performance.

Building upon above observations, we specially propose \textbf{APPO}, the \textbf{A}ttention-guided \textbf{P}erception \textbf{P}olicy \textbf{O}ptimization algorithm that enhances model's fine-grained perception through reasoning.
%%增加一部分说明为什么分为2步，或者为什么这么思考
% Since sparse perception-based outcome rewards are ineffective to provide fine-grained reward signals.
% In fact, existing popular RL methods struggle to effectively enhance model's fine-grained perception. This difficulty mainly arises because sparse outcome rewards hardly provide fine-grained guidance signals, and annotating fine-grained perceptual information is costly.
Indeed, existing popular RL methods struggle to effectively enhance model's fine-grained perception capabilities, mainly because sparse outcome rewards fail to provide sufficient fine-grained guidance signals, and fine-grained perception annotations are costly.
Therefore, to bridge these gaps, our APPO algorithm amis to establish fine-grained reward signals directly from sparse outcome rewards without relying on expensive annotation or additional reward models.
% leveraged to establish token-level fine-grained reward signals from sparse outcome rewards.
% Naturally, the attention weights of response tokens to video frames represent the model's fine-grained perception for the video, and those reasoning trajectories with higher rewards typically focus on potentially crucial video frames.
% Therefore, those response tokens can be leveraged to establish token-level fine-grained reward signals from sparse outcome rewards.
To be specific, APPO algorithm primarily consists of two core steps:
\textbf{1) Transform sparse outcome rewards into dense, frame-level guidance signals.}
% Essentially, the most intrinsic representation of perception within models is the attention scores on video frames. Therefore, it can be leveraged to identify potentially important frames from those reasoning trajectories with high outcome rewards. These frames are considered necessary to be focused on by paths with lower rewards. 
Those reasoning trajectories with higher rewards are used to identify potentially important frames. These frames are considered necessary to be focused on by paths with lower rewards.
\textbf{2) Compute token-level fine-grained rewards for intra-group perception tokens.} 
Tokens from different paths that focus on the same frames (called \emph{intra-group perception tokens}) are assigned to the same group. The discrepancy among these tokens is measered by $KL$ divergence, prioritizing learning tokens from high-reward paths, while suppressing those from low-reward paths, resulting in the token-level fine-grained rewards.
% ------
Experimental results on diverse video benchmarks and models with different scales demonstrate APPO consistently outperforms existing algorithms, such as GRPO and DAPO.
% Our work holds promise for future applications in a broader range of multimodal perception-based reinforcement learning paradigm.

In summary, our contributions are as follows:
\begin{itemize}
    \item Building upon divide-and-conquer strategy and extensive experiments, we quantify the impact of perception and reasoning on overall performance, and found that enhancing perception could yield more remarkable performance improvements compared to enhancing reasoning.
    \item To enhance model's fine-grained perception ability during reasoning, we propose APPO algorithm that encourages model to prioritize learning of intra-group perception tokens by leveraging token-level fine-grained rewards.
    \item Experimental results on diverse video benchmarks and models with different scales consistently demonstrate the effectiveness of APPO. Our work provides a promising approach to effectively enhance model’s perception abilities through reasoning in a low-cost manner, serving diverse scenarios and demands.
\end{itemize}

\section{Related Works}

\subsection{Large Language Model Reasoning}
Recent advances in Reinforcement Learning (RL) have significantly enhanced the reasoning capabilities of Large Language Models (LLMs)~\cite{jaech2024openai, guo2025deepseek, yu2025dapo, zheng2025group}. The GRPO (Group Relative Policy Optimization) proposed by DeepSeek-R1~\cite{guo2025deepseek} demonstrated that 
pure reinforcement learning can significantly enhance the robust reasoning capabilities of the model based on verifiable reward signals.
To address several key issues faced by GRPO algorithm, such as entropy collapse, reward noise, and training instability, Qiying \emph{et al.}~\cite{yu2025dapo} proposed the DAPO algorithm. 
substantially improves model's reasoning capabilities by designing verifiable reward signals, including structured thought formats and final answer accuracy. 
The DAPO~\cite{yu2025dapo}.
Further, Chujie \emph{et al.}~\cite{zheng2025group} introduced GSPO to resolve the stability challenges in the RL training of large Mixture-of-Experts (MoE) models.
The design and optimization of these policy optimization algorithms are tailored for purely textual domains. Similarly, the APPO algorithm proposed in this work is specially designed for video reasoning tasks.

\subsection{Video Reinforcement Fine-tuning}
Inspired by the success of reinforcement learning in text-based tasks, many studies~\cite{wang2025time, feng2025video, park2025deepvideo, dang2025reinforcing, chen2025exploring, ouyang2025spacer} have applied the learning paradigm to vision tasks, achieving significant improvements. 
Time-R1~\cite{wang2025time}, Space-R~\cite{ouyang2025spacer} and Video-R1~\cite{feng2025video} \emph{etc.} design proprietary reward signals for specific types of tasks, such as bounding box and time IoU to enhance the temporal and spatial reasoning capabilities of models.
To make the reasoning process of models more reliable, some studies~\cite{cao2025ground, zheng2025deepeyes, zhang2025thinking} utilize the inherent grounding capabilities of models as a tool to select the necessary data from images or videos.
While the learning paradigms used in text-based tasks have achieved decent results in these visual tasks, they fail to effectively enhance the model's perceptual abilities, thus the performance of the model remains limited.
Differently, our work starts from the perspective of video reasoning, experimentally observes and analyzes the importance of perception abilities, and specifically proposes optimization strategies to enhance model's fine-grained perception capabilities during the reasoning process.

\subsection{Perception-aware Reinforcement Learning}
Recent works~\cite{xia2025visionary,xiao2025advancing} have also recognized the importance of perception.
Visionary-R1~\cite{xia2025visionary} and Perception-R1~\cite{xiao2025advancing} introduced additional rewards that either directly assess perception quality or require the model to explicitly perform captioning before reasoning.
While promising, these strategies often impose a rigid separation between perception and reasoning, rather than enabling joint learning of both. They also rely on additional large neural-based reward models, resulting in significant computational overhead and limitations imposed by the reward model’s capacity. Our APPO algorithm could joinly improve fine-grained perception and reasoning without relying on additional reward models.

% can we jointly improve perception and reasoning in multimodal RLVR algorithms

\begin{figure*}[!t]
     \centering
     \includegraphics[width=0.95\textwidth]{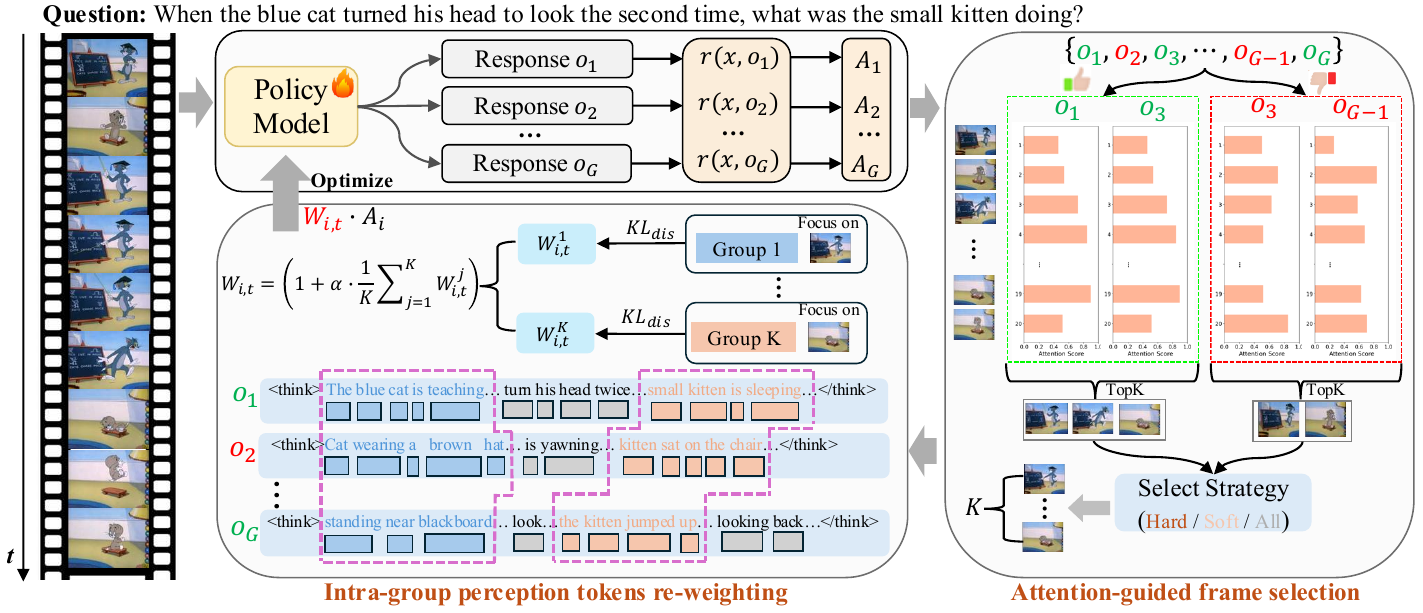}
     \vspace{-1em}
     \caption{
     The overview of APPO algorithm, which primarily consists of two core steps: attention-guided frame selection and intra-group perception tokens re-weighting.
     % The policy model generates a set of $G$ responses, and then a simple rule-based verifier is used to compute reward scores $r(x, o_i)$ and advantage $A_i$ for each response. 
     Firstly, a group of $G$ responses are divided into two sets based on the reward scores, and the final target frames are selected from these two sets based on attention weights. Building upon these frames, those tokens from different responses focusing on the same frame are assigned into a group and optimized with token-level weights.
     }
     \label{fig:framework}
     \vspace{-1.25em}
\end{figure*}

\section{Methodology}
In this section, we first introduce post-training methods for reinforcement fine-tuning in subsection~\ref{bg}, such as GRPO. Then, we propose APPO, the Attention-guided Perception Policy Optimization algorithm that enhances fine-grained perception ability while reasoning.
% we first clarify the background of video reasoning tasks in subsection~\ref{bg}, and discuss the potential issues when applying the policy optimization algorithms in language tasks to video reasoning tasks. Then, in subsection~\ref{appo}, we provide a detailed introduction to our APPO algorithm.

% associated with the application of text-based methods such as GRPO in this scenario subsection 3.1. Then, in subsection 3.2, we provide a detailed introduction to the APPO algorithm.

%  in subsection 3.1 and discuss the potential issues that may arise when applying text-based methods, such as GRPO, to video reasoning scenarios.

\subsection{Preliminary}
\label{bg}
\noindent\textbf{Task Formulation.} Following a typical setting, a training sample $x$ from dataset $\mathcal{D}$ contains video input $\mathcal{V}=\{I_{i} \in \mathbb{R}^{H \times W \times C}\}_{i=1}^{T}$, where $H, W, C$ represent the height, width and channels respectively, $T$ is the number of video frames, question $q$, and ground truth answer $a$.
% A simple rule-based verifier is used to compute rewards for each rollout during training.
We do not rely on any existing chain-of-thought data and initiate RL training directly without cold-start SFT.

\noindent\textbf{GRPO.} For an input sample $x = \{\mathcal{V}, q,a\}$, a group $G$ candidate responses $o = \{o_1, o_2, \cdots, o_G\}$ are sampled from old policy model $\pi_{\theta}^{\text{old}}$. A simple rule-based verifier is used to compute reward scores $\{r_1, r_2, \cdots, r_G\}$ for these responses. To quantify the relative quality of each response, the rewards are standardized within the group:
\begin{equation}
A_i \ = \ \frac{r_i - \mu}{\sigma}, \quad \mu  \ = \ \text{mean}(\{r_i\}), \quad \sigma \ = \ \text{std}(\{r_i\}) \ ,
\end{equation}
where $A_i$ is advantage calculated based on the relative reward within the group.
% Specifically,by assuming that the policy undergoes relatively small
% updates, i.e., ri,t(θ) ∈ (1 − ϵ, 1 + ϵ), the objective can be simplified as:
Then the policy model $\pi_\theta$ is optimized with the following objective (we omit the clip term for simplicity):
\begin{equation}
\mathcal{L}_{\text{GRPO}} \ = \ \mathbb{E}_{o \sim \pi_\theta^{\text{old}}} \left[
\frac{1}{N} r_{i,t}(\theta) \cdot A_i
- \beta \cdot \text{KL}[\pi_\theta || \pi_{\text{ref}}] \right] \ ,
\end{equation}
where $\frac{1}{N} = \frac{1}{G}\sum_{i=1}^G \frac{1}{|o_i|}\sum_{t=1}^{|o_i|}, \ r_{i,t}(\theta) = \frac{\pi_\theta(o_{i,t} | q, o_{i<t})}{\pi_{\theta}^{old}(o_{i,t}| q, o_{i<t})}$.
% Since the reward values are calculated at the sample level, they are essentially independent of tokens. However, the reality is that each response contains some perceptual tokens that focus on specific content of the video. we adopt a token-level policy gradient objective inspired by DAPO, and remove the KL penalty to enable more flexible optimization dynamics.
\subsection{Attention-guided Perception Policy Optimization (APPO)}
\label{appo}
To enhance the model's fine-grained perception capabilities during reasoning, we propose \textbf{APPO}, the \textbf{A}ttention-guided \textbf{P}erception \textbf{P}olicy \textbf{O}ptimization algorithm. 
The key idea behind APPO is to encourage the model to prioritize learning those intra-group tokens that focus on the same potentially important frame (intra-group perception tokens).
Fig.~\ref{fig:framework} shows an illustrative overview of the algorithm, which primarily consists of two core steps: attention-guided frame selection and intra-group perception tokens re-weighting. Next, we formally introduce these two key components as follows.

\noindent\textbf{Attention-guided Frame Selection.} Essentially, the most intrinsic representation of model's perception is the attention scores on video frames.
% Only when the model has focused on a particular frame, it could output relevant information. 
Those responses with higher rewards are more likely to focus on the correct video frames. 
Conversely, those with lower rewards are likely to suffer from missing or not focusing on these frames.
Therefore, this discrepancy can be used to identify potentially important video frames, thus transforming sparse outcome rewards into dense, frame-level guidance signals. 

To be specific, as shown in right part of Fig.~\ref{fig:framework}, the $G$ responses are first divided into two sets based on the reward scores, namely:
\begin{equation}
S_1 = \{ o_i \mid r_i \geq \tau \}, \quad S_2 = \{ o_i \mid r_i < \tau \},
\label{eq:divide-g}
\end{equation}
where $\tau$ is a reward threshold.
% Then, we calculate the video frames that are primarily focused on from each response. Specifically, for the $j_{th}$ token in response $o_i$, we first obtain its attention scores for $t_{th}$ frame $Attn_{j, t}^{i}$.
Then we track the attention weights from response tokens to visual tokens. For the $h$-th layer, let $a^{(h)}_{jv}$ represents the attention weight from the $j$-th response token to the $v$-th visual token. The average attention weight from the $j$-th response token to the $t$-th frame $f_{t}$ is given by:
\begin{equation}
Attn(j, f_{\mathrm{t}}) = 
% \frac{
%     \sum_{h} \sum_{v \in f_{\mathrm{t}}} a_{jv}^{(h)}
% }{
%     \sum_{h} \sum_{v \in f_{\mathrm{t}}} \mathds{1}\left( a_{jv}^{(h)} > 0 \right)
% }.
\frac{1}{\sum_{h} \cdot |f_t| } \sum_{h} \sum_{v \in f_{\mathrm{t}}} a_{jv}^{(h)},
\label{eq-4}
\end{equation}
where $|f_t|$ represents the visual token numbers of $f_t$.
Then the attention weights of response $o_i$ towards the $t$-th frame $Attn(o_i, f_t)$ is calculated by:
\begin{align}
    \phi(i) &= \operatorname{TopK}_j \big(Attn(o_{i,j}, f_t), K_1 \big), \\
    Attn(o_i, f_t) &= \frac{1}{K_1} \sum_{j \in \phi(i)} Attn(o_{i,j}, f_t),
    \label{eq-56}
\end{align}
where $o_{i,j}$ represents the $j$-th token in response $o_i$, $\phi(i)$ is the set of token indices in $o_i$, corresponding to $K_1$ tokens with the highest attention weights to $f_t$ and $|\phi(i)| = K_1$.
Following that, the video frames that primarily focused on by response $o_i$ can be denotes as:
\begin{align}
    \psi(i) = \operatorname{TopK}_t \big( Attn(o_i, f_t), K_2 \big),
    % A_t = \frac{1}{k_1} \sum_{n \in S_t} attn(n, f_t).
    \label{eq-7}
\end{align}
where $\psi(i)$ is a set of frame indices, corresponding to $K_2$ video frames that are most focused on by $o_i$ and $|\psi(i)| = K_2$.
% For simplicity, we treat the indices as corresponding video frames.
To obtain a more comprehensive set of important video frames, we take the union of $\psi(i)$ within the sets $S_1$ and $S_2$, respectively:
\begin{align}
    \psi^{S_1} = \bigcup_{o_i \in S_1} \psi(i), \quad \psi^{S_2} = \bigcup_{o_i \in S_2} \psi(i),
    \label{eq-8}
\end{align}
where $\psi^{S_1}$ and $\psi^{s_2}$ are considered the video frames most focused on by high-reward and low-reward responses.

The final selected frames $\psi'$ are selected from these two sets through three available strategies: 1) Hard. Strictly select target frames based on the difference between $\psi^{S_1}$ and $\psi^{S_2}$, that is: $\psi' = \psi^{S_1} \setminus \psi^{S_2}$;
2) Soft. To comprehensively promote the learning of frame-level fine-grained perception, use the $\psi^{S_1}$ as the target frames: $\psi' = \psi^{S_1}$;
3) All. Consider all frames focused on by both sets as target frames, namely $\psi' = \psi^{S_1} \cup \psi^{S_2}$.
Ultimately, these selected frames $\psi'$ are used to optimize those intra-group perception tokens, as described in the following.

\noindent\textbf{Intra-group Perception Tokens Re-weighting.}
For each frame in $\psi'$, there will be several tokens from $G$ responses focusing on it, also called intra-group perception tokens, which represent the model's fine-grained perception towards the same video content.
% Our goal is to model the importance of these tokens within a group, enabling the policy model to distinguish informative tokens better and optimize more effectively. 
Our goal is to model the importance of these tokens, allowing the policy model to prioritize the learning of tokens in higher rewards responses while suppressing those in lower rewards responses.
Inspired by recent works~\cite{bigelow2024forking, lin2024critical}, which demonstrate that key reasoning tokens can be identified based on token-level distributional differences, we argue that these intra-group perception tokens could be treated as crucial fine-grained tokens, and they should be given different learning intensities based on information differences among them.
% we argue that token positions where intra-group perception tokens exhibit higher divergence from the expected distribution are more likely to carry improtant information.

\begin{figure}[h]
     \centering
     \includegraphics[width=0.45\textwidth]{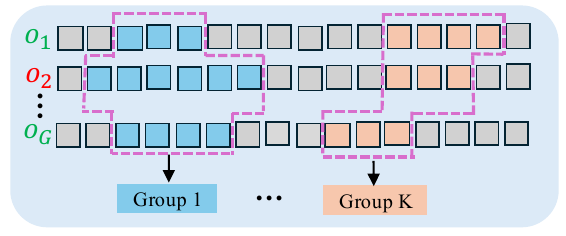}
     \vspace{-1em}
     \caption{
     The illustration of intra-group perception tokens. The tokens within the same group primarily focus on the same crucial video frame.
     }
     \label{fig:example}
     \vspace{-1.25em}
\end{figure}

In particular, we first formally define intra-group perception tokens as follows:
\begin{align}
    \Omega &= \{\Omega^{(1)}, \Omega^{(2)}, \cdots, \Omega^{(K)} \}, \\
    \Omega^{(k)} &= \big\{ \operatorname{TopK}_{o_{i,j}} \big(Attn(o_{i,j}, f_k), K_3 \big) \big \}_{i=1}^G,
    \label{eq-9-10}
\end{align}
where $\Omega$ represents $K$ groups of tokens, each group focusing on a frame $f_k$ in $\psi'$ and $K = |\psi'|$, $\Omega^{(k)}$ denotes the set of $G \times K_3$ tokens selected from each response $o_i$ that have the highest attention to a specific frame in $\psi'$.
% where $K = |\psi'|$ is the selected frame numbers, $K_3$ is the number of perception tokens selected from each response $o_i$.
A more illustrative diagram is shown in Fig.~\ref{fig:example}.
\begin{algorithm}[!t]
\small
  \caption{Attention-guided Perception Policy Optimization (APPO)}
  \label{alg:appo}
  \begin{algorithmic}[1]
  \Require
  \Statex $\pi_\theta$: Online policy model (initialized from pretrained weights)
  \Statex $\mathcal{D}$: Multimodal training dataset $\{\mathcal{V}, q, a\}$
  \Statex $\tau$: reward threshold (e.g., 0.5)
  \Statex $K_1, K_2, K_3$: hyper-parameters for TopK selection (e.g., 15, 5, 64)
  \Statex $\alpha$: hyper-parameter to control the scaling of token weights

  \Procedure{Training}{$\pi_\theta, \mathcal{D}, T$}
  \For{$t \gets 1$ to $T$}
    \For{each multimodal input $x$ in $\mathcal{D}$}
        \State \textbf{Step 1: Generation \& Rewards Computation}
        
        \State Generate $G$ responses: $\{o_i\}_{i=1}^G \sim \pi_\theta(\cdot|x)$
        
        \State Compute format and accuracy rewards: $\{r_i\}_{i=1}^G$
        
        \State Divide $G$ responses into two sets $S_1$ and $S_2$ based on format rewards in Eq.~\ref{eq:divide-g}

        \State \textbf{Step 2: Frame Selection based on Attention}
        
        \State Compute attention weights to video frame in Eq.~\ref{eq-4}
        
        \State Compute video frame indices that are most focused by $S_1$ and $S_2$ in Eq.~\ref{eq-56}~\ref{eq-7}~\ref{eq-8}

        \State Employ one select strategy to get $\psi'$

        \State \textbf{Step 3: Perception Tokens Re-weighting}

        \State Measure token discrepancy $D^{(k)}$ using $KL$ divergence in Eq.~\ref{eq-11}

        \State Normalize the divergence measurements using min-max normalization: $\mathcal{D}^{(k)} = \frac{D^{(k)} - D^{(k)}_{min}}{D^{(k)}_{max} - D^{(k)}_{min}}$

        \State Average final token weights $\mathcal{W}$ in Eq.~\ref{eq-12}

        \State Compute APPO loss in Eq.~\ref{eq-13} to optimize $\pi_\theta$

    \EndFor
        
   \EndFor
   \State \Return optimized policy $\pi_{\theta}$
\EndProcedure

  \end{algorithmic}
\end{algorithm}
% \begin{align}
%      G^k = \operatorname{TopK}\big(Attn(o_{i,j}, f_t), K_3 \big)
% \end{align}
% In particular, we introduce the token importance weight $\omega_{i,j}$ to quantify the information content of $j$-th token position among intra-group perception tokens. 
To measure the token discrepancy within $\Omega^{(k)}$, the Kullback-Leibler (KL) divergence is used to compute the probability distribition differences, which can be formalized as:
\begin{align}
    D^{(k)} = \sum_{i=1}^{G} D_{\mathrm{KL}}\left( p(\Omega^{(k)}_{i,j}) \Big\| \mathbb{E}[\Omega^{(k)}_j] \right),
    \label{eq-11}
\end{align}
where $\mathbb{E}[\Omega^{(k)}_j]$ is computed by averaging the proability distribution of each response within $\Omega^{(k)}$, and the $D^{(k)}$ represents the importance weight of each token within $\Omega^{(k)}$.
We normalize $D^k$ using min-max normalization to ensure the numerical stability.
% To this end, the $D^{(k)}$ represents the importance weight of each token within $\Omega^{(k)}$.
% \begin{align}
%     \mathcal{D}_j = \frac{D_j - D_{min}}{D_{max} - D_{min}},
% \end{align}
% To ensure comparability, the raw divergence scores are normalized, mapping them to a standard range while preserving their relative differences.
The final token importance weight is averaged by $K$ groups:
\begin{align}
    \mathcal{W} = 1 + \alpha \cdot \frac{1}{K} \sum_{k=1}^K D^{(k)},
    \label{eq-12}
\end{align}
where $\alpha$ is a hyperparameter that controls the scaling of token weights.
% Assume $P_k$ represents a group of perception tokens focusing on the $k$-th frame in $\psi$.
% Finally, this difference, multiplied by the advantage of each path, produces a token-level fine-grained perceptual reward:

To this end, the resulting weights $\mathcal{W}$ enable fine-grained perception-aware optimization by modulating the learning of token-level perception signals.
Following DAPO, we remove the KL divergence constraint to enhance the model's exploratory capabilities, and normalize the number of outputs $o_i$ to balance their contributions to the overall loss.
Therefore, the policy model is optimized with the following objective (we omit the clip term for simplicity):
\begin{equation}
\mathcal{L}_{\text{APPO}} \ = \ \mathbb{E}_{o \sim \pi_\theta^{\text{old}}} \left[
\frac{1}{\sum_{i=1}^G |o_i|} \sum_{i=1}^G \sum_{t=1}^{|o_i|} r_{i,t}(\theta) \cdot \textcolor{red}{\mathcal{W}} \cdot A_i \right] \ .
\label{eq-13}
\end{equation}

% where $\frac{1}{N} = \frac{1}{G}\sum_{i=1}^G \frac{1}{|o_i|}\sum_{t=1}^{|o_i|}$.
Above process is shown in Algorithm~\ref{alg:appo}. More details can be found in \emph{supplementary materials}.

\begin{table*}[h]\small
\centering
\caption{The comparison results across SFT, GRPO, DAPO and our APPO based on diverse video benchmarks and Qwen2.5-VL-3/7B models. All experiments are conducted on the same dataset and experimental settings to ensure a fair comparison. For evaluation, each frame is processed at a resolution of $224 \times 224$. The video is sampled at $1fps$ while maintaining the maximum of $60$ frames.}
\vspace{-1em}
\addtolength\tabcolsep{-2.4pt} 
\resizebox{0.98\linewidth}{!}{
\begin{tabular}{c|cccc|cc|c|c|c|c}
\toprule
\multicolumn{1}{c|}{\multirow{2}{*}{\textbf{Method}}} & \multicolumn{4}{c|}{\textbf{SEED-Bench-R1}} & \multicolumn{2}{c|}{\textbf{NExTGQA}} & \multicolumn{1}{c|}{\multirow{2}{*}{\textbf{Perception Test}}} & \multicolumn{1}{c|}{\multirow{2}{*}{\textbf{VSI-Bench}}} & \multicolumn{1}{c|}{\multirow{2}{*}{\textbf{MVBench}}} & \multicolumn{1}{c}{\multirow{2}{*}{\textbf{NExT-QA}}} \\ 

\multicolumn{1}{c|}{} & L1 (In-Dist.) & L2 (OOD) & L3 (OOD) & Avg. & mIoU & Acc@QA & \multicolumn{1}{c|}{} & \multicolumn{1}{c|}{} & \multicolumn{1}{c|}{} & \multicolumn{1}{c}{} \\

\midrule
\rowcolor{gray!15}\multicolumn{11}{c}{\textbf{Qwen2.5-VL-3B}} \\
% GPT-4V & - & 70.5 & 55.8 & 53.5 & 59.9 & & & & \\
Base Model & 28.2 & 29.4 & 27.0 & 28.0 & 10.1 & 40.3 & 42.9 & 29.2 & 50.8 & 59.4 \\
+ SFT & 32.6 & 33.7 & 28.6 & 31.6 & 10.5 & 63.7 & 58.2 & 32.9 & 57.1 & 72.4 \\
+ GRPO & 35.3 & 35.7 & 31.0 & 34.0 & 10.3 & 70.7 & 62.1 & 34.8 & 61.5 & 74.2 \\
+ DAPO & \underline{36.6} & \underline{37.5} & \underline{31.8} & \underline{35.3} & \underline{10.6} & \underline{70.9} & \underline{62.5} & \underline{36.7} & \textbf{62.8} & \underline{76.1} \\
\rowcolor{ModelGreen}\textbf{+ APPO (Ours)} & \textbf{37.5} & \textbf{39.1} & \textbf{35.0} & \textbf{37.2} & \textbf{11.1} & \textbf{71.2} & \textbf{63.1} & \textbf{38.2} & \underline{62.4} & \textbf{76.4} \\ 
\midrule

\rowcolor{gray!15}\multicolumn{11}{c}{\textbf{Qwen2.5-VL-7B}} \\
% GPT-4V & - & 70.5 & 55.8 & 53.5 & 59.9 & & & & \\
Base Model & 29.1 & 32.7 & 29.6 & 30.5 & 13.5 & 49.9 & 55.2 & 27.8 & 56.4 & 67.6 \\
+ SFT & 40.2 & 42.8 & 43.9 & 42.3 & 28.4 & 72.1 & 60.2 & 31.8 & 58.2 & 72.4 \\
+ GRPO & 49.0 & 49.7 & 48.2 & 48.9 & 32.0 & \underline{75.1} & 66.0 & 35.9 & 63.2 & 78.4 \\
+ DAPO & \underline{50.0} & \underline{51.3} & \underline{48.7} & \underline{50.0} & \underline{32.4} & 75.0 & \underline{66.4} & \underline{36.6} & \underline{64.2} & \textbf{79.9} \\
\midrule
% \rowcolor{gray!15}\multicolumn{10}{c}{\textbf{Video Reasoning Models}} \\
% LLaMA-VID &  & & & & & & & 41.9 & \\
% Video-R1-7B (16 Frames) & - & - & - & - & - & - & 35.8 & 63.9 & - \\
% Video-R1 & - & - & - & - & - & - & 35.8 & 63.9 & - \\
% VideoRFT & - & - & - & - & - & - & 36.8 & 61.1 & - \\
\rowcolor{ModelGreen}\textbf{+ APPO (Ours)} & \textbf{50.5} & \textbf{51.6} & \textbf{49.3} & \textbf{50.5} & \textbf{32.9} & \textbf{75.8} & \textbf{66.9} & \textbf{36.9} & \textbf{64.6} & \underline{79.6} \\ 
% \midrule

\bottomrule
\end{tabular}
}
\vspace{-0.5em}
\label{tab:main} 
\end{table*}

\begin{table*}[!t]\small
\centering
\caption{The comparison results with video reasoning models on diverse video benchmarks. 
Our APPO is trained on a subset of $34K$ from Video-R1-$260K$ dataset.
All methods employ the same base model Qwen2.5-VL-7B for zero-shot evaluation on these benchmarks. For evaluation, each frame is processed at a resolution of $224 \times 224$. The video is sampled at $1fps$ with the maximum of $30$ frames.}
\vspace{-1em}
\addtolength\tabcolsep{-2.4pt} 
\resizebox{0.98\linewidth}{!}{
\begin{tabular}{c|c|c|cccc|cc|c|c|c|c}
\toprule
\multicolumn{1}{c|}{\multirow{2}{*}{\textbf{Method}}} & \multicolumn{1}{c|}{\multirow{2}{*}{\textbf{Size}}} & \multicolumn{1}{c|}{\multirow{2}{*}{\textbf{Training Data}}} & \multicolumn{4}{c|}{\textbf{SEED-Bench-R1}} & \multicolumn{2}{c|}{\textbf{NExTGQA}} & \multicolumn{1}{c|}{\multirow{2}{*}{\textbf{Perception Test}}} & \multicolumn{1}{c|}{\multirow{2}{*}{\textbf{VSI-Bench}}} & \multicolumn{1}{c|}{\multirow{2}{*}{\textbf{MVBench}}} & \multicolumn{1}{c}{\multirow{2}{*}{\textbf{NExT-QA}}} \\ 

\multicolumn{1}{c|}{} & \multicolumn{1}{c|}{} & \multicolumn{1}{c|}{} & L1 (In-Dist.) & L2 (OOD) & L3 (OOD) & Avg. & mIoU & Acc@QA & \multicolumn{1}{c|}{} & \multicolumn{1}{c|}{} & \multicolumn{1}{c|}{} & \multicolumn{1}{c}{} \\

\midrule
% \rowcolor{gray!15}\multicolumn{10}{c}{\textbf{Qwen2.5-VL-3B}} \\
% % GPT-4V & - & 70.5 & 55.8 & 53.5 & 59.9 & & & & \\
% Base Model & 28.2 & 29.4 & 27.0 &  & & 42.9 & & & \\
% + SFT &  & & & & & & & & \\
% + GRPO & 35.3 & 35.7 & 31.0 & 11.0 & 70.7 & 62.1 & & & \\
% + DAPO & 36.6 & 37.5 & 31.8 & 11.0 & 70.9 & 62.9 & & & \\
% \rowcolor{ModelGreen}\textbf{APPO (Ours)} & 37.5 & 39.1 & 35.0 & 11.0 & 71.2 & 63.1 & & & \\ 
% \midrule

\rowcolor{gray!15}\multicolumn{13}{c}{\textbf{Video Reasoning Models}} \\
TW-GRPO~\cite{dang2025reinforcing} & 7B & $1K$ & 30.2 & 29.0 & 25.6  & 28.3 & 17.7 & 53.2 & 54.9 & 25.1 & 52.9 & 70.2 \\ 
VideoChat-R1~\cite{li2025videochat} & 7B  & $18K$ & \underline{33.3} & \underline{34.4} & 27.8 & \underline{31.8} & \textbf{22.0} & 63.2 & 57.7 & 19.9 & 55.8 & 74.5 \\
GRPO-CARE~\cite{grpogrpo} & 7B & $260K$ & 29.9 & 28.3 & 26.0 & 28.1 & \underline{18.0} & 51.5 & 57.1 & 24.0 & 54.6 &  66.4 \\
% TinyLLaVA-Video-R1~\cite{zhang2025tinyllava} & 3B & & & & & & & & & & \\
Video-R1~\cite{feng2025video} & 7B & $260K$ & 30.9 & 32.3 & \underline{28.8} & 30.7 & 14.9 & 64.7 & 57.0 & \textbf{35.8} & \underline{63.9} & 76.5 \\
VideoRFT~\cite{wang2025videorft} & 7B & $310K$ & 32.4 & 32.5 & 27.7 & 30.9 & 14.4 & \underline{72.0} & \underline{60.4} & 30.3 & 59.0 & \underline{78.3} \\
% DeepVideo-R1~\cite{park2025deepvideo} & & & & & & & & & \\
% Time-R1~\cite{wang2025time} & & & & & & & & & \\
\midrule

\rowcolor{ModelGreen}\textbf{APPO (Ours)} & 7B & $34K$ & \textbf{35.4} & \textbf{40.0} & \textbf{34.7} & \textbf{36.1} & 16.9 & \textbf{76.3} & \textbf{64.7} & \underline{32.7} & \textbf{64.6} & \textbf{79.2} \\ 
% \midrule

\bottomrule
\end{tabular}
}
\vspace{-0.5em}
\label{tab:comp} 
\end{table*}

\section{Experiments}

\subsection{Experimental Setup}
\noindent\textbf{Evaluation.}
To validate the effectiveness of our proposed APPO, we conduct evaluations on diverse video benchmarks, including: (1) video reasoning tasks (\emph{e.g.}, SEED-Bench-R1~\cite{chen2025exploring}, VSI-Bench~\cite{yang2025thinking}, Perception Test~\cite{patraucean2023perception}); (2) fine-grained spatial-temporal video reasoning tasks (NExT-GQA~\cite{xiao2024can}) and (3) general video understanding tasks (\emph{e.g.},  MVBench~\cite{li2024mvbench}, NExT-QA~\cite{xiao2021next}).

\noindent\textbf{Implementation details.}
We perform direct RL training from Qwen2.5-VL-3B and 7B models, comparing the standard GRPO and DAPO baseline with our proposed APPO.
% We employ Qwen2.5-VL-3B and Qwen2.5-VL-7B for the experiments. 
We use the accuracy and format rewards to train the model on $6K$ SEED-Bench-R1 and $6K$ Perception Test multi-choice training subset for $1$ epoch, respectively. The IoU reward is added to train models on $3K$ NExT-GQA validation set for $2$ epochs. The weights of three rewards are $0.9$, $0.1$, and $0.9$. 
For zero-shot evaluation on VSI-Bench, MVBench and NExT-QA, we select a subset of $34K$ training data from Video-R1-$260K$ RL dataset and train the model for $1$ epoch.
The rollout numbers for per input sample in set to $8$. 
APPO uses the same clip-higher strategy as DAPO and dynamic sampling strategy was not used.
Training batch size is $16$ and learning rate is fixed as $1e-6$. Each frame is processed at a resolution of $224 \times 224$ while maintaining a maximum of $30$ frames for $3$B model and $16$ frames for $7$B model during training.
We only use accuracy reward scores to group all responses, and the reward threshold $\tau$ in Eq.~\ref{eq:divide-g} is set to $0.5$. 
% The impact of differenet hyperparameters such as $K_1, K_2, K_3$ and $\alpha$ is demonstrated in the ablation experiments.
More training details are provided in the \emph{supplementary materials}.

\subsection{Main Results}
To verify the effectiveness of APPO, we compared with SFT, GRPO and DAPO based on the same dataset and base model (Qwen2.5-VL-3/7B) to ensure a fair evaluation. 
The main results are shown in Table.~\ref{tab:main}.

\noindent\textbf{RL generally enhances performance compared to SFT.}
It can be found that RL methods such as GRPO, DAPO, and APPO have effectively improved model performance compared to traditional SFT, both in video understanding and reasoning benchmarks, indicating the significant advancements in recent RL explorations.

\noindent\textbf{Overall performance improvement compared to GRPO and DAPO on diverse video benchmarks.}
APPO consistently outperforms GRPO and DAPO on video reasoning and general understanding benchmarks.
Specifically, for video reasoning tasks, such as SEED-Bench-R1 and VSI-Bench, APPO demonstrated a significant performance improvement ($1.5\% \sim 3.2\%$ on 3B model and $ 0.3\% \sim 1.6\%$ on 7B model).
On general video understanding dataset, including MVBench and NExT-QA, APPO achieved comparable performance. For fine-grained spatiotemporal perception tasks, \emph{e.g.} NExT-GQA, APPO shows overall performance improvements on mIoU and accuracy metrics.

% Specifically, for video reasoning tasks, APPO achieved a performance improvement of on SEED-Bench-R1 and VSI-Bench, respectively, compared to DAPO.
% Specifically, on the SEED-Bench-R1 benchmark, APPO achieved overall performance improvements of $3.2\%$ and $1.9\%$ on the 3B model, $1.6\%$ and $0.5\%$ improvements on the 7B model.
% Specifically, on the 3B model, APPO achieved overall performance improvements of $1.9\%$ and $1.5\%$ compared to DAPO on the SEED-Bench-R1 and VSI-Bench benchmarks, respectively. 

\noindent\textbf{APPO demonstrates greater advantages in enhancing fine-grained perception capabilities.}
It is worth noting that the performance improvement on the 7B model is not as remarkable as on the 3B model on SEED-Bench-R1 benchmark ($0.5\%$ \emph{vs.} $1.9\%$ compared with DAPO, $1.6\%$ \emph{vs.} $3.2\%$ compared with GRPO). We attribute this to the weaker perception capabilities of the 3B model, where the advantage of APPO in enhancing fine-grained perception is more prominently displayed.
This phenomenon is also reflected in the NExT-GQA dataset, where the mIoU metric primarily measures the model's fine-grained spatiotemporal perception capabilities. It can be observed that the GRPO and DAPO algorithms have limited performance improvements in grounding the correct frames ($0.2\%$ and $0.4\%$ improvements on 3B model). In comparison, APPO could achieve improvements of $1.0\%$ on 3B model, and the consistent conclusion is also drawn for the 7B model.

\noindent\textbf{Stronger generalization capabilities.}
For OOD test data (Level-2 and Level-3) on SEED-Bench-R1 benchmark, where the testing data comes from Ego4D~\cite{grauman2022ego4d}, including cross-environment tasks, APPO achieved a more remarkable improvement over DAPO ($1.6\%$ and $3.2\%$ on 3B model), which is significantly greater than performance improvement on Level-1 in-distribution test data ($0.9\%$ on 3B model), demonstrating stronger generalization capabilities.

% APPO shows consistent overall improvements ($0.9\% \sim 4\%$) with identical training dataset, rollout space and reward design compared to the GRPO and DAPO algorithms.  
%% 增加reward 曲线对比，以及熵之类的
% \noindent\textbf{More significant improvements on OOD reasoning.}
% The performance gains are more pronounced on OOD test set (Level-2 and Level-3 on SEED-Bench-R1 benchmark).

% \noindent\textbf{Improved performance on fine-grainded perception.}
% Furthermore, for fine-grained spatial-temporal video reasoning tasks, APPO also achieved higher mIoU and accuracy.

% \input{tabs/tab_comp}

%% 增强video-r1训练在其他benchmark上的对比结果
\begin{figure*}[!t]
    \centering
    \begin{subfigure}[b]{0.32\textwidth}
        \centering
        \includegraphics[width=\textwidth]{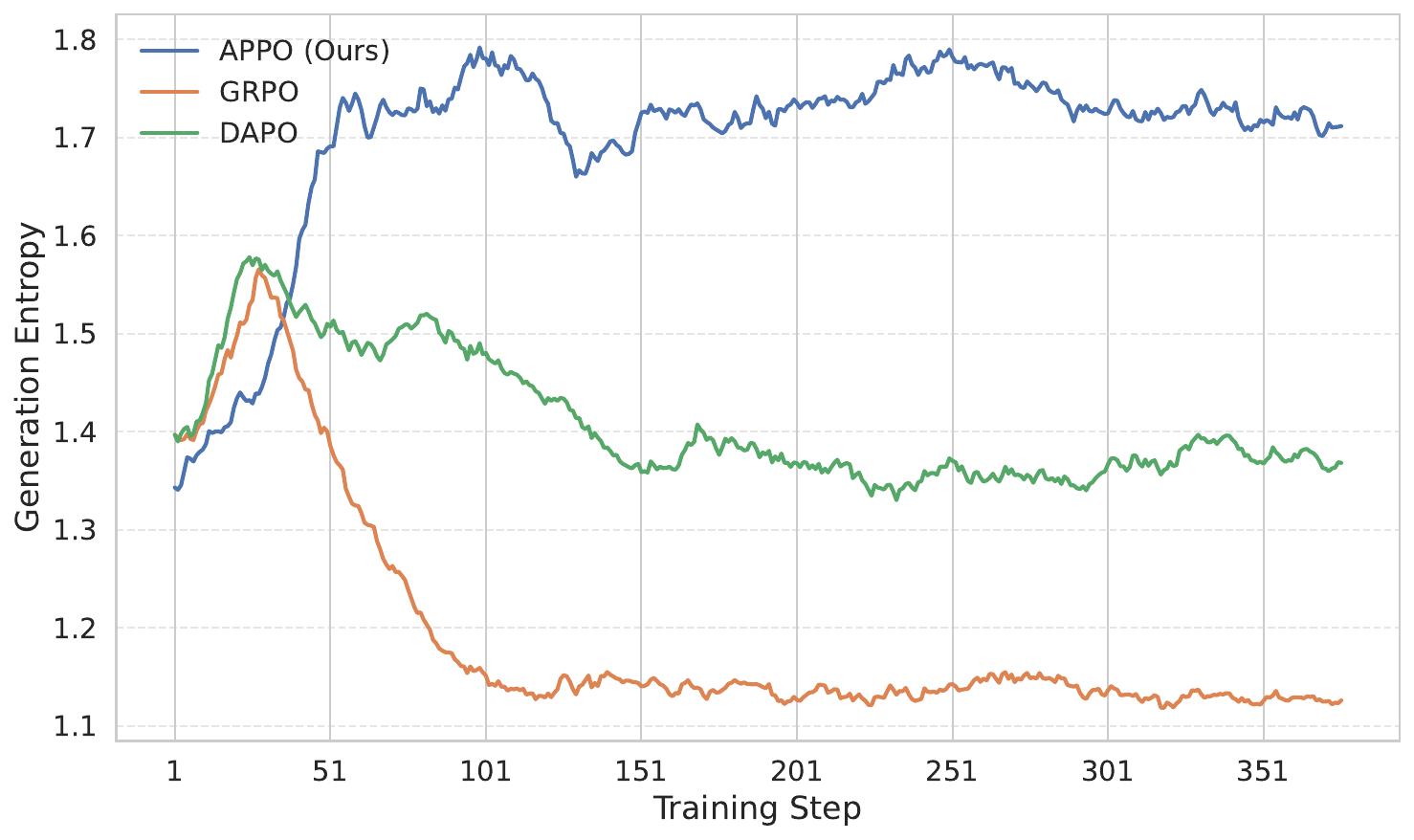}
        \caption{The comparison of generation entropy.}
        \label{fig:entropy}
    \end{subfigure}
    \begin{subfigure}[b]{0.32\textwidth}
        \centering
        \includegraphics[width=\textwidth]{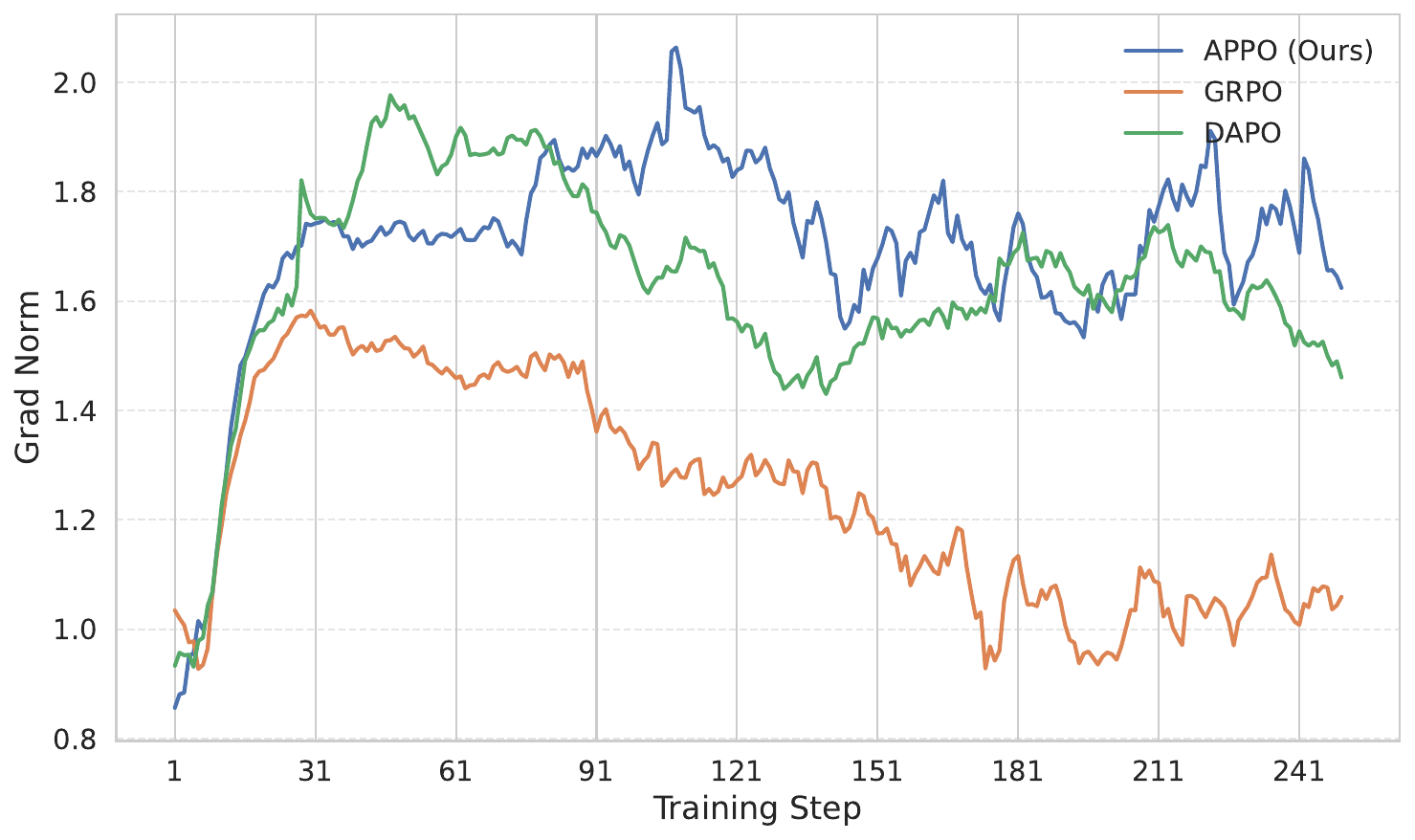}
        \caption{The comparison of grad norm.}
       \label{fig:grad-norm}
    \end{subfigure}
    \begin{subfigure}[b]{0.32\textwidth}
        \centering
        \includegraphics[width=\textwidth]{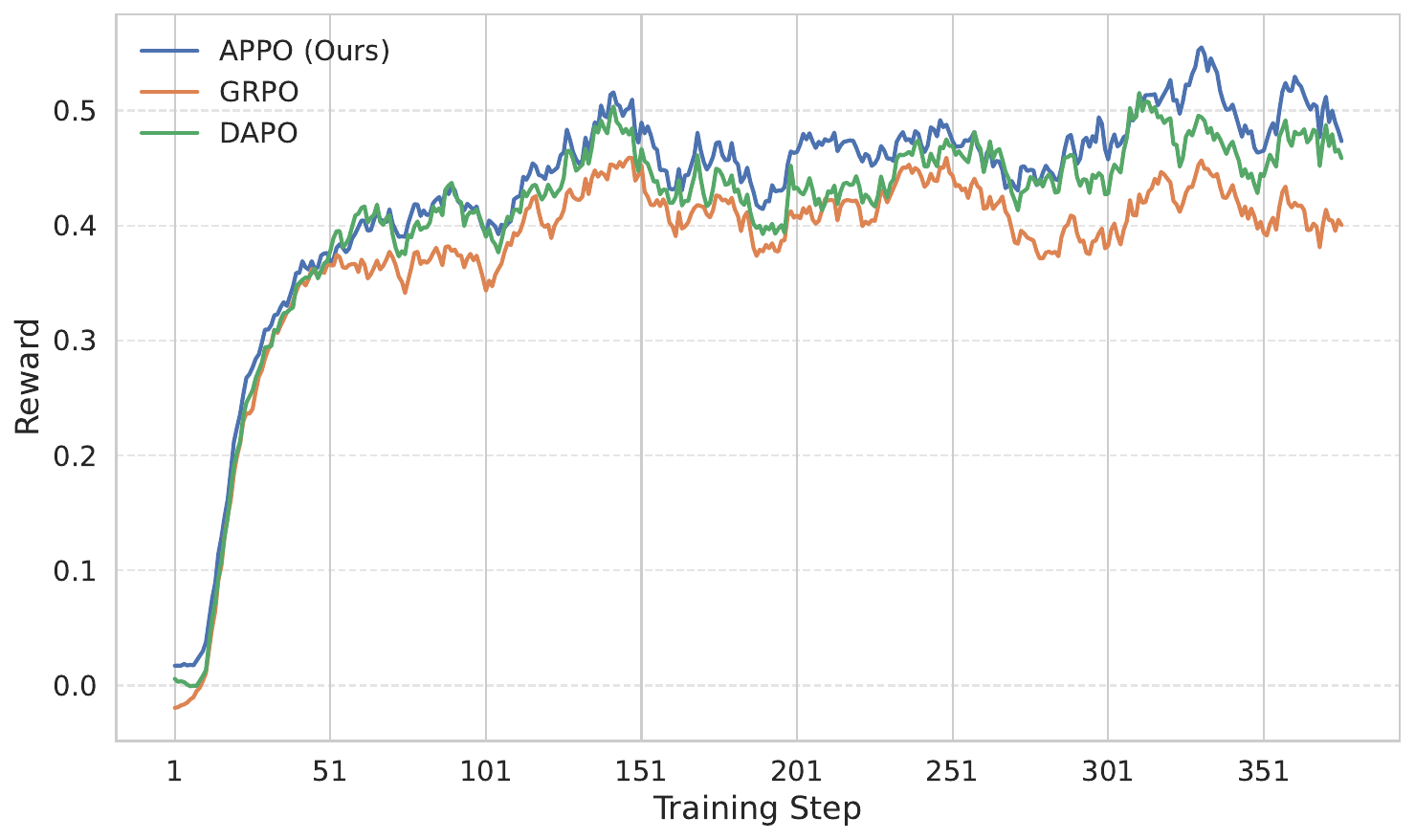}
        \caption{The comparison of reward score.}
        \label{fig:reward}
    \end{subfigure}
    \vspace{-1em}
    \caption{The comparison of generation entropy, grad norm and reward scores during training process with GRPO and DAPO.}
    \label{fig:analysis}
    \vspace{-0.4em}
\end{figure*}

\begin{figure}[h]
     \centering
     \includegraphics[width=0.4\textwidth]{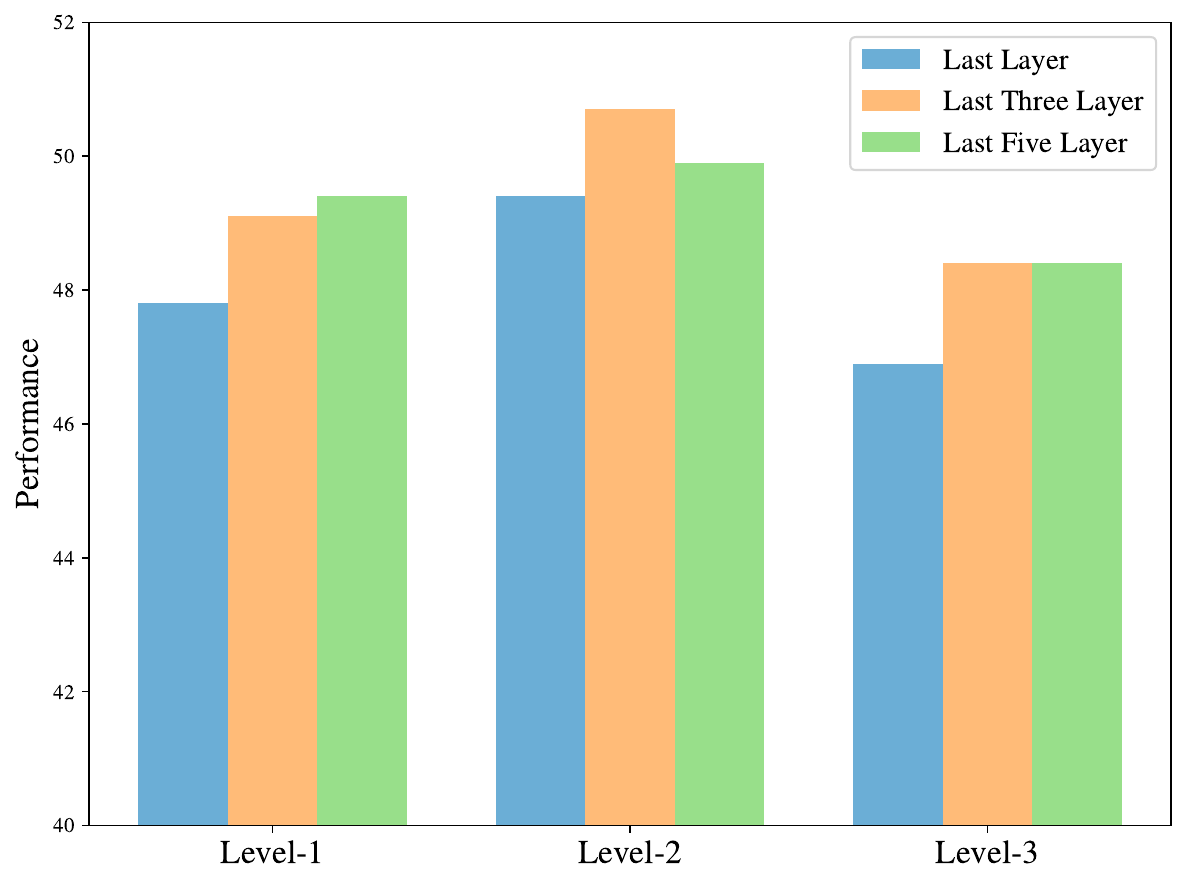}
     \vspace{-1em}
     \caption{
     The ablation results for different attention layers.
     }
     \label{fig:ab-attn-layer}
     \vspace{-1.25em}
\end{figure}

\subsection{Comparison with other video reasoning models}
To verify the advantages of our APPO algorithm compared to existing works, we selected a subset of $34K$ training data from the Video-R1-$260K$ RL dataset and trained the Qwen2.5-VL-7B model using APPO algorithm. Detailed composition of training data is provided in the \emph{supplementary materials}.
Table.~\ref{tab:comp} shows the comparison results of APPO with existing video reasoning models, including TW-GRPO~\cite{dang2025reinforcing}, GRPO-CARE~\cite{grpogrpo}, Video-R1~\cite{feng2025video}, VideoRFT~\cite{wang2025videorft} and VideoChat-R1~\cite{li2025videochat}. We also present the training data scales for these models. All methods employ the same base model Qwen2.5-VL-7B for zero-shot evaluation on these benchmarks.

It can be observed that while APPO is trained on only a small dataset of $34K$, it demonstrates overall  performance improvements compared to other methods trained on larger-scale datasets.
% indicating the data efficiency brought by enhanced fine-grained perception capabilities.
In particular, the APPO achieves the superior results on the SEED-Bench-R1, Perception Test, VSI-Bench, MVBench, and NExT-QA benchmarks, yielding a performance improvement of $0.7\% \sim 5.2\%$, which demonstrates the remarkable advantages of APPO in enhancing fine-grained perception during reasoning.
For NExT-GQA benchamrk, the training data for Video-Chat-R1 includes temporal grounding tasks, which significantly enhances its temporal grounding ability. Compared to other models, APPO achieves comparable performance, particularly excelling in multi-choice accuracy ($76.3\%$).
The overall performance improvement achieved by APPO demonstrates the effectiveness of enhancing fine-grained perception.

\subsection{Comparison and analysis of training process}
To further understand the training process of APPO, we compared and analyzed the dynamic changes of generation entropy, gradient norm, and reward scores with GRPO and DAPO during training process. As shown in Fig.~\ref{fig:analysis}, it can be observed that during the training process, APPO exhibits higher generation entropy and grad norm, indicating the model has a larger exploration space. We attribute this phenomenon to the gains brought about by optimizing those intra-group perception tokens.
The reward score curve in Fig.~\ref{fig:reward} also indicates APPO has higher rewards, resulting in superior performance.

% \begin{figure*}[htbp]
%     \centering
%     \begin{subfigure}[b]{0.32\textwidth}
%         \centering
%         \includegraphics[width=\textwidth]{figs-pdf/entropy.pdf}
%         \caption{The comparison of generation entropy.}
%         \label{fig:entropy}
%     \end{subfigure}
%     \begin{subfigure}[b]{0.32\textwidth}
%         \centering
%         \includegraphics[width=\textwidth]{figs-pdf/grad-norm.pdf}
%         \caption{The comparison of grad norm.}
%        \label{fig:grad-norm}
%     \end{subfigure}
%     \begin{subfigure}[b]{0.32\textwidth}
%         \centering
%         \includegraphics[width=\textwidth]{figs-pdf/reward.pdf}
%         \caption{The comparison of reward score.}
%         \label{fig:reward}
%     \end{subfigure}
%     \vspace{-1em}
%     \caption{The comparison of generation entropy, grad norm and reward scores during training process with GRPO and DAPO.}
%     \label{fig:analysis}
%     \vspace{-0.4em}
% \end{figure*}

\begin{figure*}[htbp]
    \centering
    \begin{subfigure}[b]{0.24\textwidth}
        \centering
        \includegraphics[width=\textwidth]{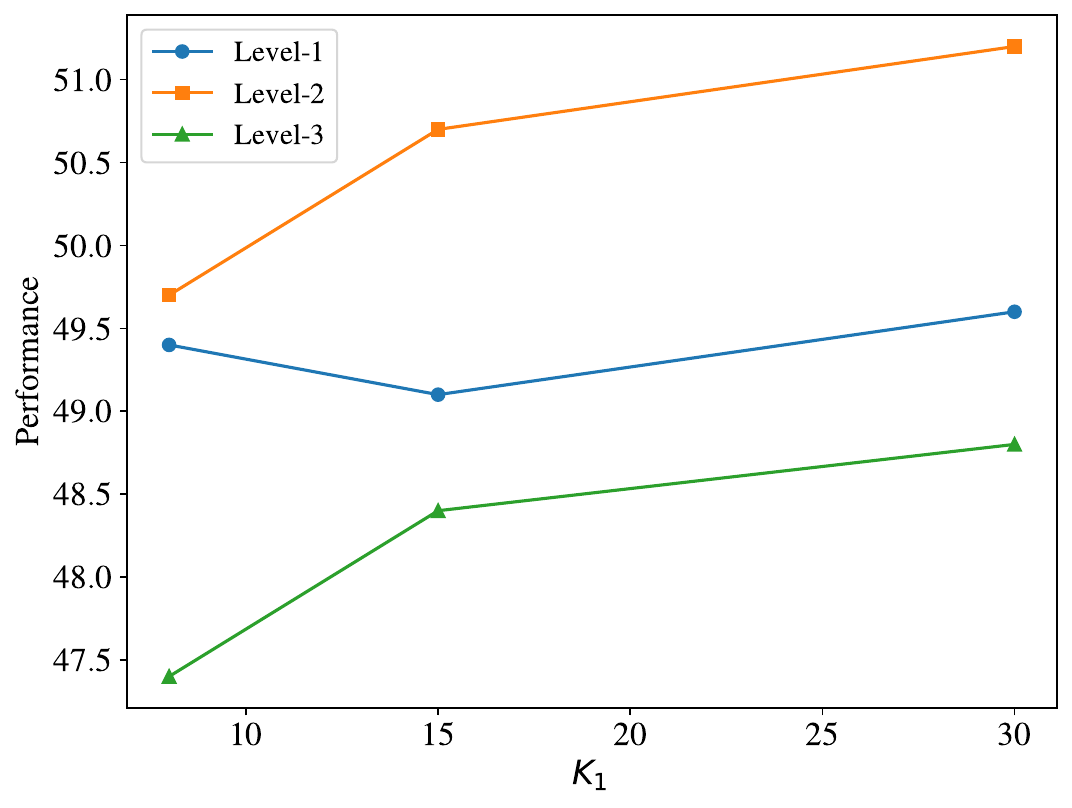}
        \caption{Ablation results for $K_1$.}
       \label{fig:ab-k1}
    \end{subfigure}
    \begin{subfigure}[b]{0.24\textwidth}
        \centering
        \includegraphics[width=\textwidth]{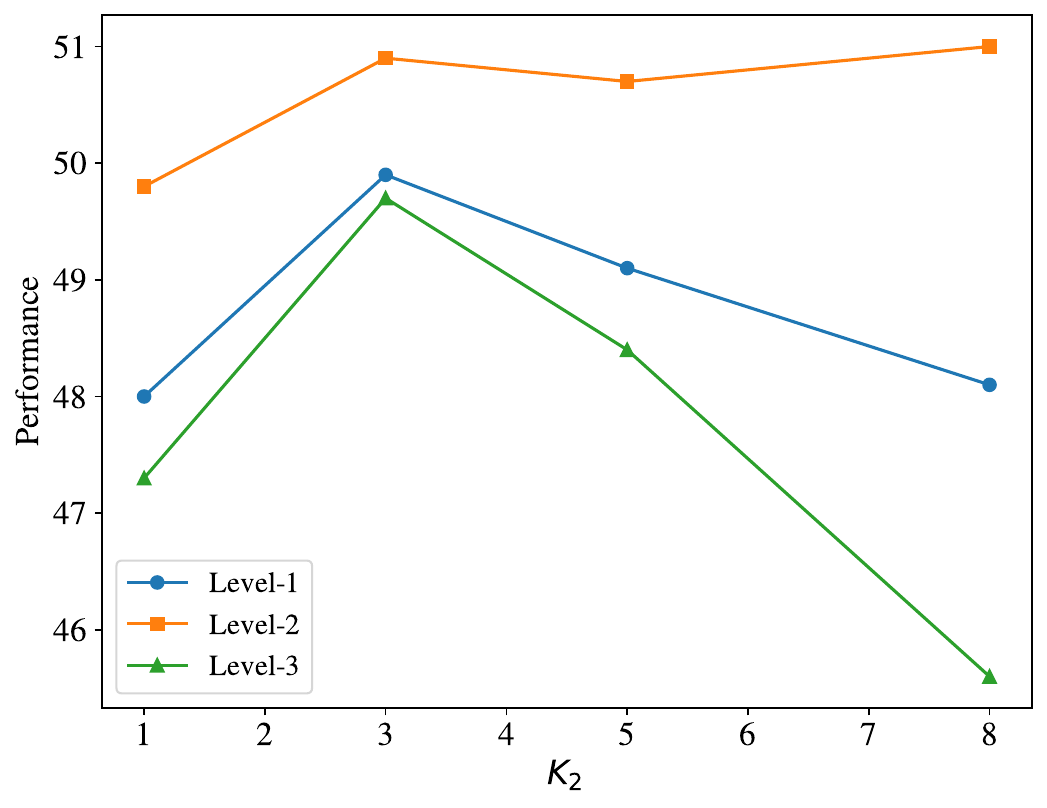}
        \caption{Ablation results for $K_2$.}
        \label{fig:ab-k2}
    \end{subfigure}
    \begin{subfigure}[b]{0.24\textwidth}
        \centering
        \includegraphics[width=\textwidth]{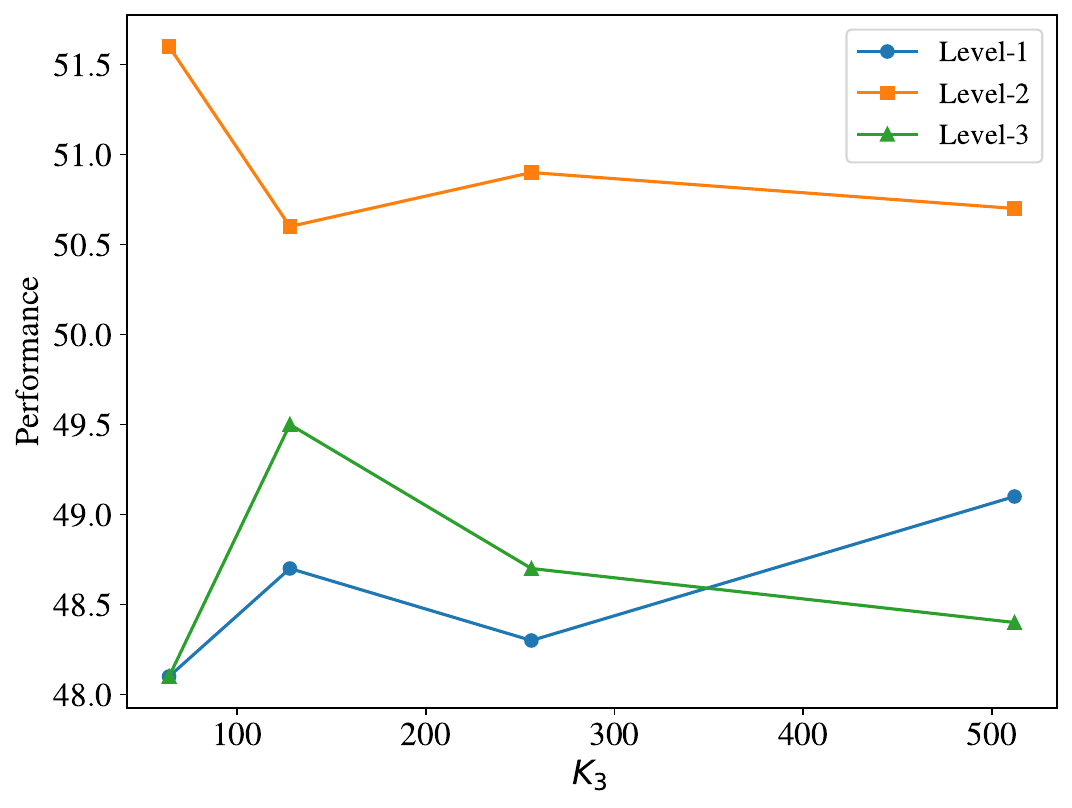}
        \caption{Ablation results for $K_3$.}
        \label{fig:ab-k3}
    \end{subfigure}
    \begin{subfigure}[b]{0.24\textwidth}
        \centering
        \includegraphics[width=\textwidth]{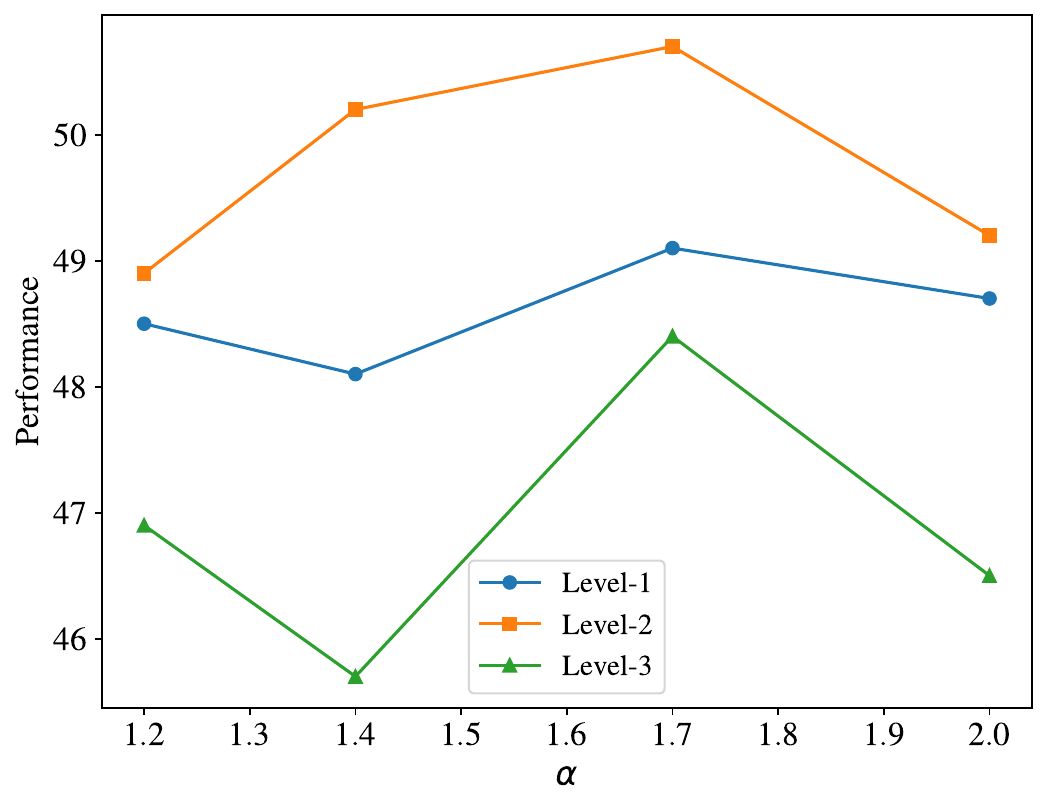}
        \caption{Ablation results for $\alpha$.}
        \label{fig:ab-alpha}
    \end{subfigure}
    \vspace{-1em}
    \caption{The ablation results for different hyperparameters in APPO based on SEED-Bench-R1 benchmark and Qwen2.5-VL-7B model.}
    \label{fig:ab}
    \vspace{-0.4em}
\end{figure*}

% \begin{figure*}[!t]
%     \centering
%     \begin{subfigure}[b]{0.24\textwidth}
%         \centering
%         \includegraphics[width=\textwidth]{CVPR2026/figs/ab_alpha.png}
%         \caption{Ablation results for $\alpha$.}
%         \label{fig:ab-alpha}
%     \end{subfigure}
%     \begin{subfigure}[b]{0.24\textwidth}
%         \centering
%         \includegraphics[width=\textwidth]{CVPR2026/figs/ab_k1.png}
%         \caption{Ablation results for $K_1$.}
%         \label{fig:ab-k1}
%     \end{subfigure}
%     \begin{subfigure}[b]{0.24\textwidth}
%         \centering
%         \includegraphics[width=\textwidth]{CVPR2026/figs/ab_k2.png}
%         \caption{Ablation results for $K_2$.}
%         \label{fig:ab-k2}
%     \end{subfigure}
%     \begin{subfigure}[b]{0.24\textwidth}
%         \centering
%         \includegraphics[width=\textwidth]{CVPR2026/figs/ab_k3.png}
%         \caption{Ablation results for $K_3$.}
%         \label{fig:ab-k3}
%     \end{subfigure}
%     \caption{The ablation results for different hyper-parameters in APPO based on SEED-Bench-R1 benchmark and Qwen2.5-VL-7B model.}
%     \label{fig:ab}
%     \vspace{-1em}
% \end{figure*}

% \begin{figure}[h]
%      \centering
%      \includegraphics[width=0.4\textwidth]{figs-pdf/attn_layer_ablation.pdf}
%      \vspace{-1em}
%      \caption{
%      The ablation results for different attention layers.
%      }
%      \label{fig:ab-attn-layer}
%      \vspace{-1.25em}
% \end{figure}

\subsection{Ablation study for APPO}
To analyze impact of hyperparameters in APPO algorithm, we conducted comprehensive ablation experiments based on SEED-Bench-R1 dataset and Qwen2.5-VL-7B model as follows, mainly including: (1) $K_1$ in Eq.~\ref{eq-56}; (2) $K_2$ in Eq.~\ref{eq-7}; (3) $K_3$ in Eq.~\ref{eq-9-10}; (4) $\alpha$ in Eq.~\ref{eq-12} and (5) attention layers to compute attention weights.
More ablation experiments, such as three select strategies, are provided in \emph{supplementary materials}.

\noindent\textbf{Ablation results for $K_1, K_2$ and $K_3$.}
As shown in Fig.~\ref{fig:ab-k1}, a larger $K_1$ generally leads to overall performance improvements because it can more accurately represent the attention weights of response $o_i$ on video frames, enhancing the reliability of selected frames.
In Fig.~\ref{fig:ab-k2}, there is an optimal value for $K_2$, which is $K_2 = 3$. The possible reason is that smaller $K_2$ might lead to missing some important frames, while larger $K_2$ could increase the number of selected frames, introducing frame noise and misleading the optimization direction.
For $K_3$ in Fig.~\ref{fig:ab-k3}, larger $K_3$ means that more perception tokens are selected in each response. However, since the content of each frame is limited, too many tokens are unnecessary and could interfere with the model's optimization of truly useful perception tokens.

% \noindent\textbf{Ablation results for $K_2$.}

% \noindent\textbf{Ablation results for $K_3$.}

\noindent\textbf{Ablation results for $\alpha$.}
The $\alpha$ in Eq.~\ref{eq-12} controls the regulation strength of perception tokens. As shown in Fig.~\ref{fig:ab-alpha}, as $\alpha$ increases from $1.2$ to $2.0$, the performance across three levels first increases and then decreases, reaching a maximum when $\alpha=1.7$. Moreover, the variation in $\alpha$ has a greater impact on OOD test data (maximum performance difference $1.8\%$ on L2 and $2.7\%$ on L3) compared to in-distribution data (maximum performance difference $1\%$), indicating a close relationship between optimization strength of tokens and generalization capabilities.

\noindent\textbf{Ablation results for attention layers.}
In Eq.~\ref{eq-4}, we need to calculate the average attention scores across several layers to obtain the attention weights. To explore the impact of attention layers on performance, we compared three strategies: the last layer, the last three layers and the last five layers.
As shown in Fig.~\ref{fig:ab-attn-layer}, using only the last layer results in the worse performance, possibly due to the lack of sufficiently accurate attention information. As the number of attention layers increases from $3$ to $5$, there is a gradual improvement in performance. However, considering training efficiency, an excessive number of attention layers can consume more GPU memory. Therefore, using the last three layers might be the optimal choice.

% \noindent\textbf{ablation for filter samples}

% \noindent\textbf{dynamic sampling exp}

\section{Conclusion}
In this work, we innovatively decouple the model's perception and reasoning capabilities. Through extensive experimental observations, we found that enhancing perception abilities could bring more significant improvements compared to enhancing reasoning. Building upon this observation, to fully enhance the model's fine-grained perception abilities during reasoning process, we propose the APPO algorithm, which utilizes token-level fine-grained rewards to optimize the model. Experimental results on diverse video benchmarks and models of different scales consistently show that APPO is more suitable for video reasoning tasks compared to mainstream algorithms like GRPO and DAPO. Our work provides a promising approach to effectively enhance model's perception abilities through reasoning in a low-cost manner, serving diverse scenarios.

% \input{sec/X_suppl}
% {
%     \small
%     \bibliographystyle{ieeenat_fullname}
%     \bibliography{main}
% }

% \input{sections/introduction}
% \input{sections/relatedwork}
% \input{sections/approach}
% \input{sections/experiments}

\clearpage

\bibliographystyle{IEEEtran}
\bibliography{paper}

\clearpage

\beginappendix

% \input{sections/appendix}
% \clearpage
\setcounter{page}{1}
\appendix
\renewcommand{\thefigure}{\Alph{figure}}
\renewcommand{\thetable}{\Alph{table}}
% \maketitlesupplementary

\hypersetup{
	colorlinks=true,
	linkcolor=red,
	filecolor=blue, 
	urlcolor=red,
	citecolor=green,
}

% \tableofcontents

\section{Details of APPO}
Our APPO algorithm primarily consists of two core steps: Attention-guided Frame Selection and Intra-group Perception Tokens Re-weighting, as mentioned in Section 3.2 of main submission.
% Therefore, in this section, we primarily discuss two aspects: the rationality of frame selection and token re-weighting.
Therefore, in this section, we mainly discuss rationality and correctness of these two steps in detail.

\subsection{Attention-guided Frame Selection}
\begin{figure}[!t]
     \centering
     \includegraphics[width=0.48\textwidth]{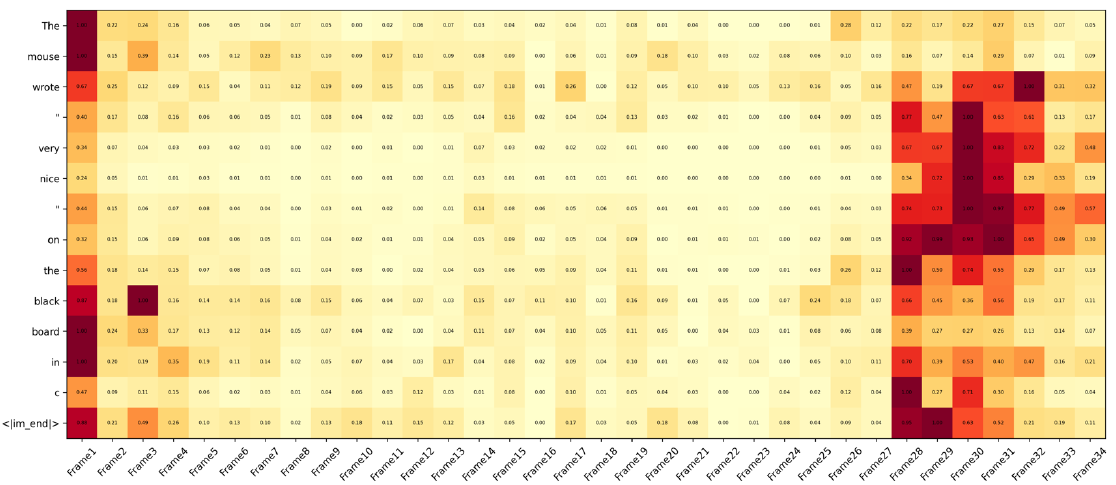}
     \vspace{-1em}
     \caption{
     The attention weights of response tokens to video frames. Question: ``What did the mouse write on the blackboard?". The answer mainly appears in the last part of the video. The x-axis represents the video frame index (total $34$ frames). The y-axis shows the response tokens. The darker the color, the higher the attention score.
     }
     \label{fig:attn-1}
     \vspace{-1.25em}
\end{figure}

\begin{figure}[!t]
     \centering
     \includegraphics[width=0.48\textwidth]{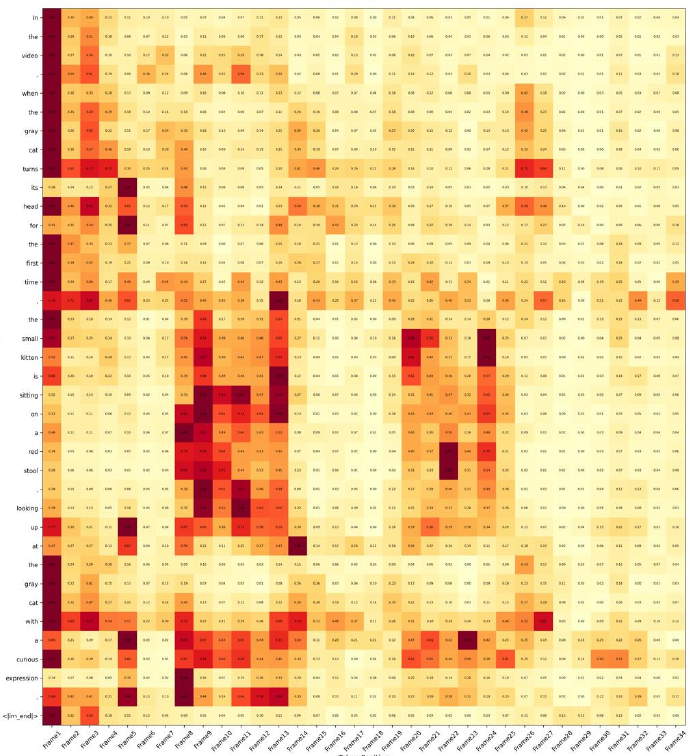}
     \vspace{-1em}
     \caption{
      The attention weights of response tokens to video frames. Question: ``When the blue cat turned his head to look the second time, what was the small kitten doing?". The answer mainly appears in the middle part of the video ($15s \sim 20s$). The x-axis represents the video frame index (total $34$ frames). The y-axis shows the response tokens. The darker the color, the higher the attention score.
     }
     \label{fig:attn-2}
     \vspace{-1.25em}
\end{figure}

To understand the limitations of MLLMs' visual perception, recent works~\cite{zhang2025mllms} have studied the attention patterns of MLLMs when answering visual questions. Inspired by these works, we also explored the attention patterns in video scenarios.
The Fig.~\ref{fig:attn-1} and Fig.~\ref{fig:attn-2} present the attention patterns of response tokens to the video frames, respectively. It can be found that: (1) For the same video, the model exhibits different patterns of attention weights on video frames as the question changes; (2) The model's incorrect answers are due to either missing crucial frames or assigning too low attention weights to these frames. 
In particular, as shown in Fig.~\ref{fig:attn-1}, the question is ``What is the mouse writing on the blackboard?", and the answer mainly appears in the last part of the video. 
% The model correctly focuses on the content of the last few frames. 
The model correctly focused on the last few frames, especially when it provided correct information such as ``very nice".
Similarly, in Fig.~\ref{fig:attn-2}, the question is ``What is the kitten doing when the blue cat turns its head for the second time?", and the answer mainly appears in the middle part of the video ($16s \sim 21s$). Clearly, the model did not pay sufficient attention to these video frames, resulting in an incorrect answer.

These empirical observations inspire us that the reasoning paths with higher rewards are more likely to focus on crucial frames.
Therefore, as mentioned in Section 3.2 of main submission, we can divide all reasoning paths into two sets based on the reward scores, and then utilize the attention differences between these two sets to obtain frame-level guiding signals.

\subsection{Intra-group Perception Tokens Re-weighting}
The main goal of this step is to determine different learning intensities for intra-group perception tokens.
Recent works~\cite{bigelow2024forking, lin2024critical} demonstrate that the key reasoning tokens can be identified based on token-level distributional differences.
Building upon this conclusion, Jisheng~\emph{et al.}~\cite{dang2025reinforcing} calculated the weights of tokens at each position for the entire sequence. 
Similarly, we argue that the intra-group perception tokens for each crucial frame could be treated as crucial fine-grained perception tokens, and they should be given different learning intensities based on information differences among them.

\begin{figure}[!t]
     \centering
     \includegraphics[width=0.48\textwidth]{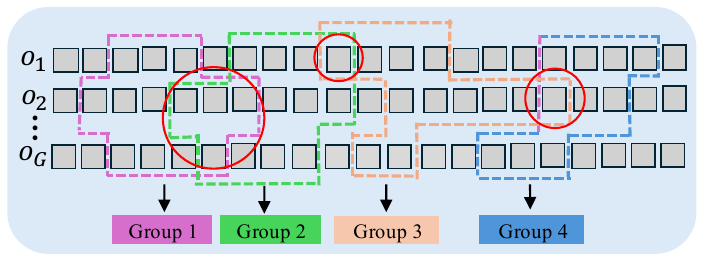}
     \vspace{-1em}
     \caption{
     The illustration of four intra-group perception tokens. The same token may attend to multiple crucial frames, as shown in the red solid circles in the figure.
     }
     \label{fig:supp-illu}
     \vspace{-1.25em}
\end{figure}

In particular, for each crucial frame $\mathcal{I}$ in $\psi'$, each path contains certain perception tokens that focus primarily on this frame. The importance weights of this group of tokens at each position can be computed as follows:
\begin{align}
    D^{(k)} = \sum_{i=1}^{G} D_{\mathrm{KL}}\left( p(\Omega^{(k)}_{i,j}) \Big\| \mathbb{E}[\Omega^{(k)}_j] \right),
    \label{eq-11}
\end{align}
where $\mathbb{E}[\Omega^{(k)}_{j} ]$ is computed by averaging the proability distribution of each response within $\Omega^{(k)}$, and the $D^{(k)}$ represents the importance weight of each token within $\Omega^{(k)}$.
Since the same token may attend to several crucial frames in $\psi'$, these tokens may correspond to multiple optimization intensities, as shown in the red solid circles in Fig.~\ref{fig:supp-illu}. 
Therefore, we additionally record the numbers of each token is optimized and the cumulative sum of the optimization intensities, then leverage the average intensity as the final token weights.

% \begin{align}
%     \mathcal{W} = 1 + \alpha \cdot \frac{1}{K} \sum_{k=1}^K D^{(k)},
%     \label{eq-12}
% \end{align}

\subsection{Relationship between answer correctness and focused frames}
To analyze relationship between answer correctness and focused frames, we conducted statistics on NExTGQA test set, which features including keyframe labels to ensure reliability of this analysis. 
Specifically, we calculated average differences in focused frames between correct and incorrect answers from the same sample, shown in Tab.~\ref{tab:re-diff}.
Approximately $20\%$ of results \emph{focus on the same video frames, regardless of correctness of final answer, making it difficult to select key frames by attention}.
\emph{In most cases ($70\% \sim 80\%$), the key frames could be distinguished based on the correctness of final answer.}
For those few special cases, it might be a meaningful direction for future exploration.

\begin{table}[!t]\small
\centering
% \vspace{-0.8em}
\caption{Differences in focused frames on NExTGQA test set.}
\vspace{-1em}
\addtolength\tabcolsep{-2.4pt} 
\resizebox{0.9\linewidth}{!}{
\begin{tabular}{c|c|c|c|c|c|c|c}
\toprule
\textbf{Attention Difference} & 0s &  0 $\sim$ 3s & 3 $\sim$ 6s & 6 $\sim$ 9s & 9 $\sim$ 12s & 12 $\sim$ 15s & $> 15s$ \\
\midrule
\textbf{Sample Nums} & 20.1\% & 23.5\% & 17.0\% & 12.6\% & 7.8\% & 6.2\% & 12.8\% \\
\bottomrule
\end{tabular}
}
\vspace{-1em}
\label{tab:re-diff} 
\end{table}

\section{Detailed Experimental Setup}
For RL training, we limit the maximum number of response tokens to 512 for training efficiency. Each training step processes 16 input samples, with 8 rollouts per sample. Rollout sampling uses a temperature of 1.0 and top-p of 1.0. Specifically, for GRPO, we use a KL divergence coefficient of 0.01. For DAPO, the clip higher is set to 0.28. For our APPO, to obtain attention weights, we set the model's attention implementation to ``eager" during sampling and average the weights across all attention heads. We use the official Hugging Face TRL library for training.

For evaluation, we use the vLLM~\cite{kwon2023efficient} inference engine to accelerate, with a sampling temperature of 0.1, top-p of 0.001, top-k of 1, and a repetition penalty of 1.05. For accuracy calculation, we use regular expressions to match the predicted results with the ground truth. For mIoU calculation, we strictly compute the intersection and union between the predicted results and the labels.

Detailed hyperparameters duraing training and evaluation are shown in Tab.~\ref{tab:supp-settings}.

\begin{table}[!t]
\centering
\caption{The hyperparameters used during training and evaluation.}
\vspace{-1em}
\begin{tabular}{l|l}
\toprule
\textbf{Hyper-parameters}         & \textbf{Value}      \\
\midrule
\rowcolor{gray!15}\multicolumn{2}{c}{\textbf{Training}} \\
Batch size                        & 16                  \\
Gradient Accumulation Steps       & 1                   \\
Warmup                            & False               \\
Rollout Numbers                   & 8                  \\
Rollout Temperature               & 1.0                 \\
Rollout Top-P                     & 1.0                 \\
Freeze Vision Encoder             & True                \\
KL divergence coefficient         & $1 \times 10^{-2}$  \\
Learning rate                     & $1 \times 10^{-6}$  \\
GPUs                              & 16                  \\
Optimizer                         & AdamW               \\
Training Framework                & TRL                 \\
\midrule
\rowcolor{gray!15}\multicolumn{2}{c}{\textbf{Evaluation}} \\
Inference Engine                  & vLLM                \\
Temperature                       & 0.1                \\
Top-P                             & 0.001                 \\
Top-K                             & 1                   \\
Repetition Penalty                & 1.05                \\
\bottomrule
\end{tabular}
\label{tab:supp-settings}
\end{table}

\section{Training Data Composition}
\begin{table}[!t]
\centering
\caption{The detailed composition of $34K$ training subset selected from Video-R1-$260K$ RL training data.}
\vspace{-1em}
\begin{tabular}{cc}
\toprule
\textbf{Data Resource}                  &            \textbf{Sample Nums.} \\
\midrule
LLaVA-Video                      &            $\sim 5K$   \\
STAR                              &            $\sim 11K$   \\
Perception Test                   &            $\sim 6K$    \\
NExT-QA                        &              $\sim 5K$   \\
CLEVRER                        &               $\sim 6K$   \\
% \midrule
% Total                       &                 $\sim 34K$  \\
\bottomrule
\end{tabular}
\label{tab:supp-train-set}
\end{table}

To validate the effectiveness of our APPO compared to GRPO and DAPO, we collected subsets from existing dataset for RL training, including: $6K$ SEED-Bench-R1 training subset, $6K$ Perception Test multiple-choice subset, $3K$ NExT-GQA validation subset.
To compare APPO with other video reasoning models, we selected $34K$ subset from Video-R1-$260K$ RL training data, as summaried in Tab.~\ref{tab:supp-train-set}.

\section{Additional Experiment Results}
\subsection{Ablation results}
To investigate the impact of the three selection strategies on the APPO algorithm, we conducted ablation experiments on SEED-Bench-R1 benchmark, as shown in Tab.~\ref{tab:select}.
It can be observed that the Soft selection strategy performs the best, as it can fully promote fine-grained perception learning in both the high reward path set and the low reward path set.

\begin{table}[t]\small
\centering
\caption{Ablation results for different select strategies on SEED-Bench-R1 Benchmark.}
\vspace{-1em}
\addtolength\tabcolsep{-2.4pt}
\resizebox{0.95\linewidth}{!}{
\begin{tabular}{c|ccc}
\toprule
\textbf{Strategy} & \textbf{L1 (In-Dist.)} & \textbf{L2 (OOD)} & \textbf{L3 (OOD)} \\
\midrule
% \rowcolor{gray!15}\multicolumn{4}{c}{\textbf{Qwen2.5-VL-3B}} \\
% \alpha=1.2 &  & &  \\
% \alpha=1.4 & &  &  \\
% \alpha=1.7 &      &      &      \\
% \alpha=2.0 &  &  &  \\
% \midrule

% \rowcolor{gray!15}\multicolumn{4}{c}{\textbf{Qwen2.5-VL-7B}} \\
Hard & 49.1 & 50.7 & 48.4 \\
% Soft & 46.6 & 48.2 & 44.7 \\
Soft & \textbf{50.5} & \textbf{51.3} & \textbf{49.3} \\
All & 50.2 & 50.9 & 47.3 \\

\bottomrule
\end{tabular}
}
% \vspace{-2mm}
% \vspace{-2mm}
\label{tab:select} 
\end{table}

\subsection{Stability and robustness}
Ablation studies in the main paper show that model performance is stable (within $0.5\%$) when hyperparameters are set within suitable ranges (\emph{e.g.}, $K_1$: $15\sim20$, $K_2$: $2\sim4$, \emph{etc.}). 
Additionally, five results with different seeds were used to calculate standard deviation (see Tab.~\ref{tab:re-std}), showing stable improvements on $3$/$7$B models.

\begin{table}[!t]\small
\centering
\caption{Standard deviation results of Tab 1.}
\vspace{-1.em}
\addtolength\tabcolsep{-2.4pt}
\resizebox{0.9\linewidth}{!}{
\begin{tabular}{c|ccc|c|c}
\toprule
\multicolumn{1}{c|}{\multirow{2}{*}{\textbf{Model}}} & \multicolumn{3}{c|}{\textbf{SEED-Bench-R1}}  & 
\multicolumn{1}{c|}{\multirow{2}{*}{\textbf{Perception Test}}} & \multicolumn{1}{c}{\multirow{2}{*}{\textbf{MVBench}}} \\

\multicolumn{1}{c|}{} & L1 (In-Dist.) & L2 (OOD) & L3 (OOD) & \multicolumn{1}{c|}{} & \multicolumn{1}{c}{} \\

\midrule

Qwen2.5-VL 3B & 0.238 & 0.319 & 0.279 & 0.301  & 0.257 \\
Qwen2.5-VL 7B & 0.103 &  0.183 & 0.201 & 0.163 &  0.095 \\

\bottomrule
\end{tabular}
}
\vspace{-1em}
\label{tab:re-std} 
\end{table}

For example, since APPO relies on the model's attention scores, this means that when sampling multiple reasoning paths, the model's attention implementation must be "eager." Additionally, existing inference acceleration frameworks, such as vLLM, have difficulty supporting the retrieval of attention scores, which limits the efficiency of the APPO algorithm.

\section{Prompt Template}
The prompt templates used for RL training on different tasks are as follows.

\begin{figure}[h]
     \centering
     \includegraphics[width=0.4\textwidth]{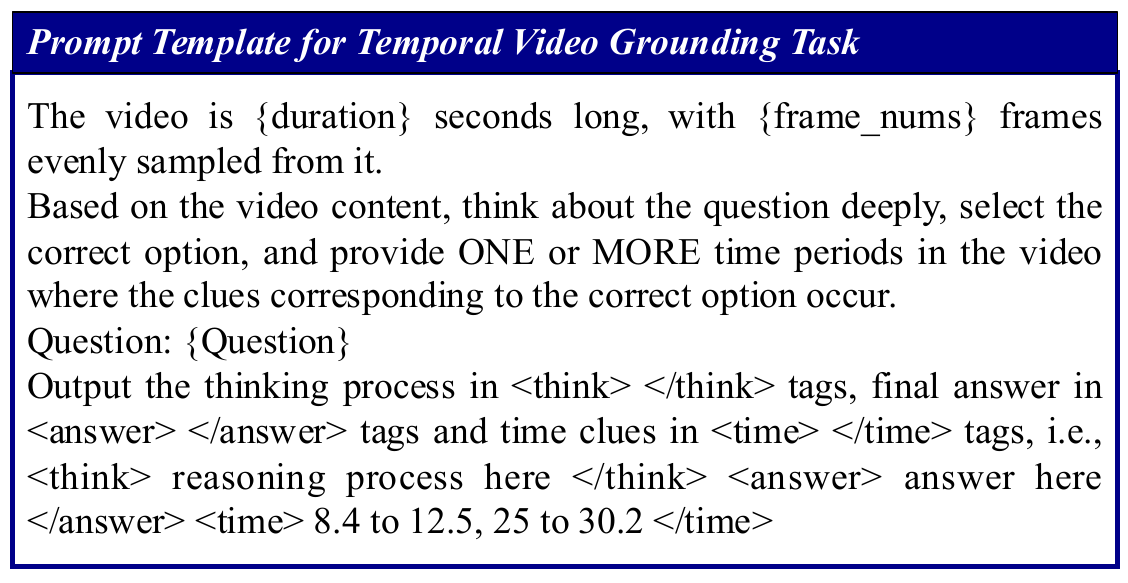}
     \vspace{-1em}
     \caption{
    The prompt template for temporal video grounding task.
     }
     \label{fig:prompt-1}
     \vspace{-1.25em}
\end{figure}

\begin{figure}[h]
     \centering
     \includegraphics[width=0.4\textwidth]{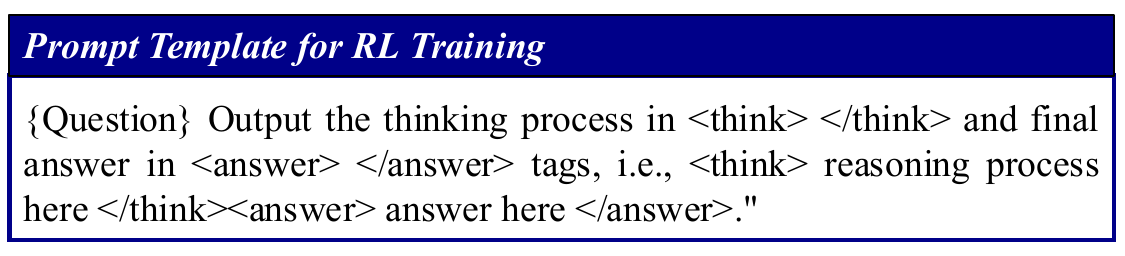}
     \vspace{-1em}
     \caption{
     The prompt template for RL training.
     }
     \label{fig:prompt-2}
     \vspace{-1.25em}
\end{figure}

\begin{table}[!t]\small
\centering
\caption{The training efficiency comparison between DAPO and our APPO. Training efficiency refers to the training samples processed per second.
}
\vspace{-1em}
\addtolength\tabcolsep{-2.4pt} 
\resizebox{0.98\linewidth}{!}{
\begin{tabular}{c|cc|c}
\toprule
\multirow{2}{*}{\textbf{Model Scale}} & \multicolumn{2}{c|}{\textbf{Training Efficiency}} & \multirow{2}{*}{\textbf{Percentage}} \\
& DAPO & APPO (Ours) & \\
\midrule
3B (30 frames) & 0.325 & 0.264 &  81\%  \\
7B (16 frames) & 0.462 & 0.429 & 93\%  \\
\bottomrule
\end{tabular}
}
\vspace{-0.5em}
\label{tab:effe} 
\end{table}

% \begin{tabular}{|c|c|c|c|}
% \hline
% Scale & DAPO Effe & APPO Effe & Percetntage \\
% \hline
% 7B & $0.462$ & $0.429$ & $93\%$ \\
% 3B & $0.325$ & $0.264$ & $81\%$ \\
% \hline
% \label{tab:effe}
% \end{tabular}

\section{Visualization Results}
\begin{figure}[!t]
  % \vspace{-1em}
  \centering
  \begin{subfigure}[b]{0.22\textwidth}
    \includegraphics[width=\textwidth]{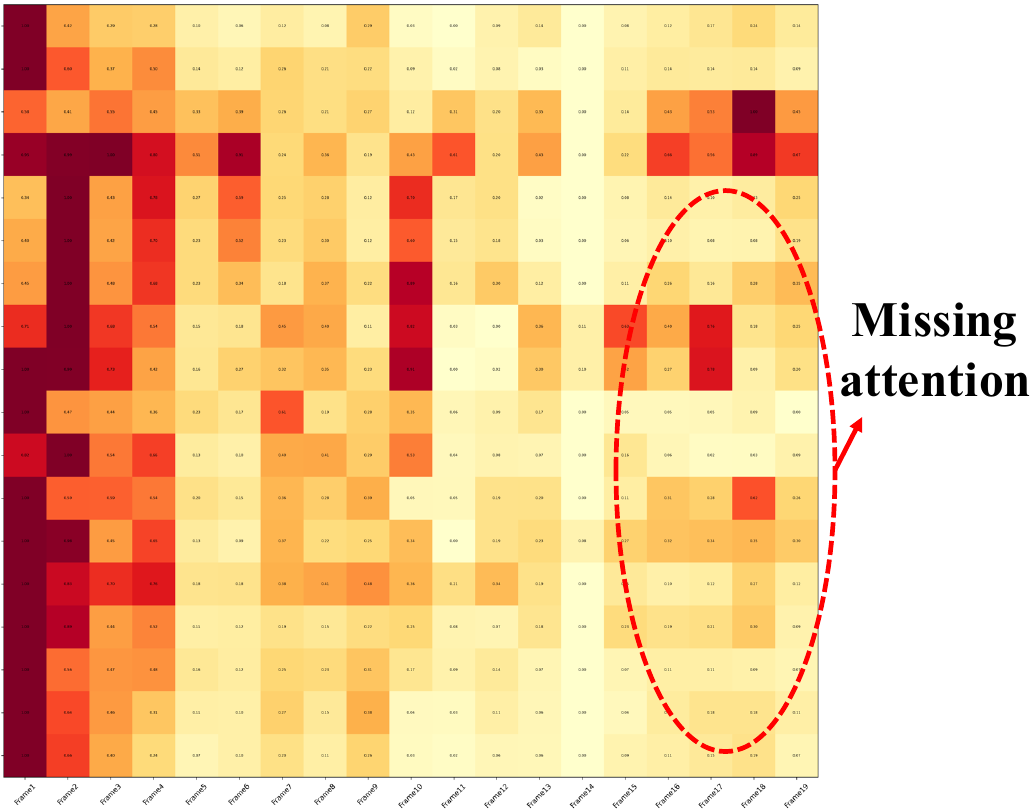}
    % \vspace{-0.2em}
    \caption{DAPO attention result.}
    \label{fig:dapo-vis}
  \end{subfigure}
  % \hfill % Optional: adjust the horizontal spacing between the subfigures
  \hspace{0.3cm}
  %\hspace{-0.05cm}
  \begin{subfigure}[b]{0.22\textwidth}
    \includegraphics[width=\textwidth]{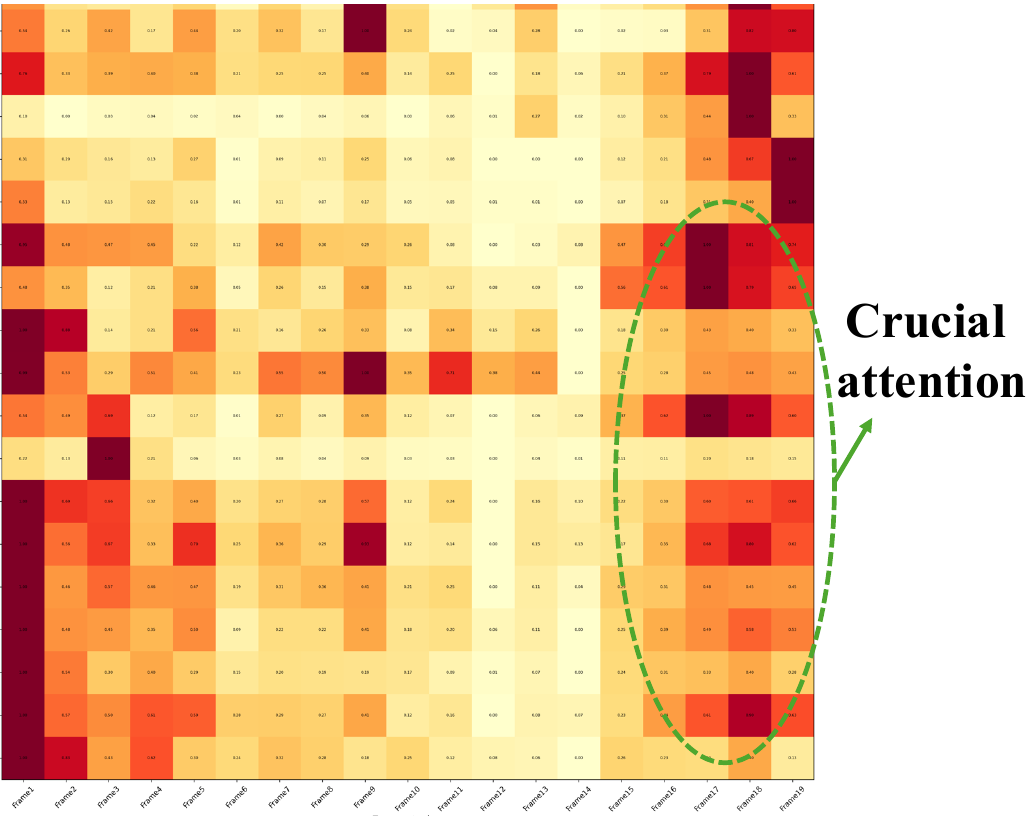}
    \caption{APPO attention result.}
    \label{fig:dapo-vis}
  \end{subfigure}
  \vspace{-1.2em}
  \caption{Attention comparison between DAPO and our APPO. \textbf{X-axis}: video frames. \textbf{Y-axis}: generated tokens during reasoning.}
  \label{fig:attn-vis}
  \vspace{-1em}
\end{figure}

\begin{figure*}[h]
     \centering
     \includegraphics[width=0.9\textwidth]{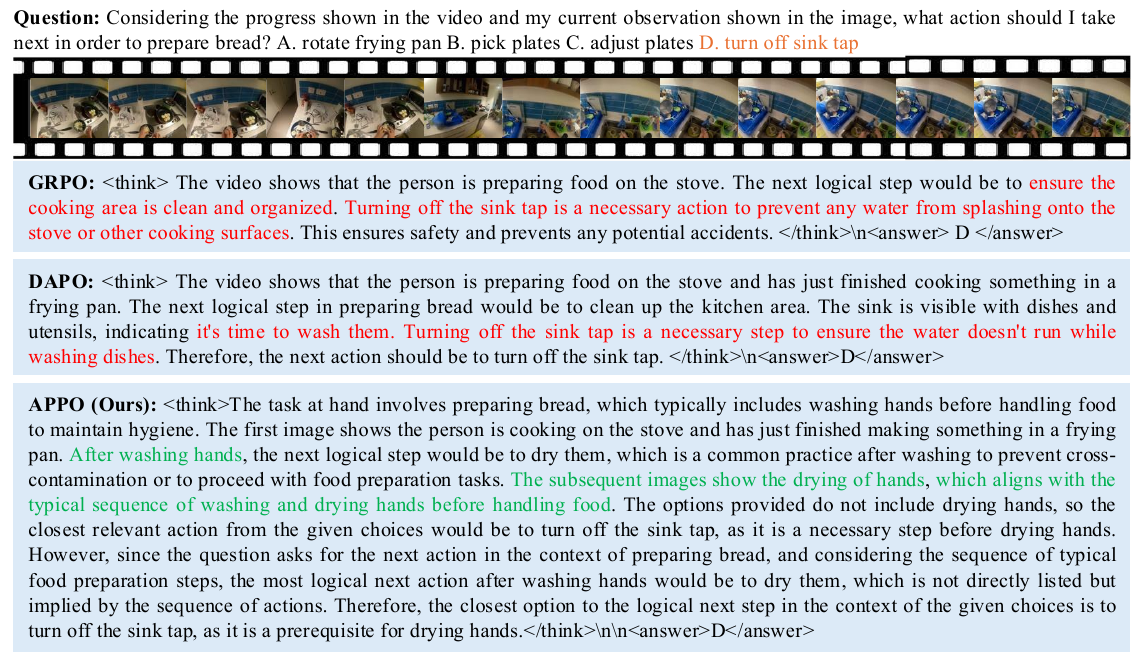}
     \vspace{-1em}
     \caption{
     The visualization example from SEED-Bench-R1 Level-3 testset. The correct option is ``D. turn off sink tap". Our APPO algorithm successfully paid attention to the critical information of handwashing, which made subsequent logical reasoning more coherent.
     }
     \label{fig:vis}
     \vspace{-1.25em}
\end{figure*}

As shown in Fig.~\ref{fig:vis}, we present visualization example from SEED-Bench-R1 Level-3 testset.
It can be found that while GRPO and DAPO algorithms correctly answered question, it is noteworthy that our APPO algorithm successfully paid attention to the critical information of \emph{handwashing}, which made the subsequent logical reasoning more coherent.
Additionally, the attention visualization results in Fig.~\ref{fig:attn-vis} demonstrate that APPO enables model to focus more on crucial frames during reasoning process, resulting in perception improvement.

\section{Limitations}
While APPO achieves superior results than GRPO and DAPO in complex video reasoning scenarios, it also has some limitations. 
For example, since APPO relies on the model's attention scores, this means that when sampling multiple reasoning paths, the model's attention implementation must be ``eager". Additionally, existing inference acceleration frameworks, such as vLLM, have difficulty supporting the output of attention scores.
However, based on our experimental statistics, the training efficiency of APPO is $81\% \sim 93\%$ that of DAPO under the same training data and model scale, as shown in Tab.~\ref{tab:effe}. 
% We look forward to researchers effectively addressing this limitation in the future. 
We will propose effective solutions to mitigate this limitation in the future works.

\end{document}